\documentclass{article}

     \usepackage[preprint]{neurips_2019}

\usepackage[utf8]{inputenc} 
\usepackage[T1]{fontenc}    
\usepackage{hyperref}       
\usepackage{url}            
\usepackage{booktabs}       
\usepackage{amsfonts}       
\usepackage{nicefrac}       
\usepackage{microtype}      
\usepackage[pdftex]{graphicx}
\usepackage{subfig}
\usepackage{wrapfig, blindtext}
\usepackage{eqnarray,amsmath} 

\usepackage{caption}

\usepackage{algorithmic}
\usepackage{algorithm}

\title{Towards Understanding Generalization in Gradient-Based Meta-Learning}

\author{%
  Simon Guiroy \\
  Mila \\
  Universit\'e de Montr\'eal\\
  \texttt{simon.guiroy@umontreal.ca} \thanks{{Correspondence to: Simon Guiroy <simon.guiroy@umontreal.ca>}} \\
  \And
  Vikas Verma \\
  Mila\\
  Aalto University, Finland\\
  \texttt{vikasverma.iitm@gmail.com} \\
  \And
  Christopher Pal \\
  Mila \\
  \'Ecole Polytechnique de Montr\'eal\\
  ElementAI, Montr\'eal\\
  \texttt{christopher.pal@polymtl.ca}  \\
}

\renewcommand\footnotemark{}
\begin{document}

\maketitle

\begin{abstract}
  In this work we study generalization of neural networks in gradient-based meta-learning by analyzing various properties of the objective landscapes. We experimentally demonstrate that as meta-training progresses, the meta-test solutions, obtained after adapting the meta-train solution of the model, to new tasks via few steps of gradient-based fine-tuning, become flatter, lower in loss, and further away from the meta-train solution. We also show that those meta-test solutions become flatter even as generalization starts to degrade, thus providing an experimental evidence against the correlation between generalization and flat minima in the paradigm of gradient-based meta-leaning. Furthermore, we provide empirical evidence that generalization to new tasks is correlated with the coherence between their adaptation trajectories in parameter space, measured by the average cosine similarity between task-specific trajectory directions, starting from a same meta-train solution. We also show that coherence of meta-test gradients, measured by the average inner product between the task-specific gradient vectors evaluated at meta-train solution, is also correlated with generalization. Based on these observations, we propose a novel regularizer for MAML and provide experimental evidence for its effectiveness.
\end{abstract}

\section{Introduction}

To address the problem of the few-shot learning, many meta-learning approaches have been proposed recently \citep{DBLP:journals/corr/FinnAL17}, \citep{sachin}, \citep{DBLP:journals/corr/abs-1810-06784}, \citep{DBLP:journals/corr/abs-1805-10123} and
\citep{DBLP:journals/corr/SnellSZ17} among others. In this work, we take steps towards understanding the characteristics of the objective landscapes of the loss functions, and their relation to generalization, in the context of gradient-based few-shot meta-learning. While we are interested in understanding the properties of optimization landscapes that are linked to generalization in gradient-based meta-learning in general, we focus our experimental work here within a setup that follows the recently proposed Model Agnostic Meta-Learning (MAML) algorithm \citep{DBLP:journals/corr/FinnAL17}. The MAML algorithm is a good candidate for studying gradient-based meta-learning because of its independence from the underlying network architecture.
\newpage

Our main insights and contributions can be summarized as follows:
\begin{enumerate}
    \item As gradient-based meta-training progresses:
    \begin{itemize} 
        \item the adapted meta-test solutions become flatter on average, while the opposite occurs when using a finetuning baseline.
        \item the adapted final solutions reach lower average support loss values, which never increases, while the opposite occurs when using a finetuning baseline.
     \end{itemize}
    \item When generalization starts to degrade due to overtraining, meta-test solutions keep getting flatter, implying that, in the context of gradient-based meta-learning, flatness of minima is not correlated with generalization to new tasks.
    \item We empirically show that generalization to new tasks is correlated with the coherence between their adaptation trajectories, measured by the average cosine similarity between trajectory directions. Also correlated with generalization is the coherence between meta-test gradients, measured by the average inner product between meta-test gradient vectors evaluated at meta-train solution.
    \item Based on this observation on coherence of adaptation trajectories , we propose a novel regularizer for gradient-based meta-learning and experimentally demonstrate its effectiveness in regularizing MAML.
\end{enumerate}

\section{Related work}\label{sec:related_works}

 There has been extensive research efforts on studying the optimization landscapes of neural networks in the standard supervised learning setup. Such work has focused on the presence of saddle points versus local minima in high dimensional landscapes \citep{DBLP:journals/corr/PascanuDGB14},\citep{DBLP:journals/corr/DauphinPGCGB14}, the role of overparametrization in generalization \citep{2016arXiv161101540F}, loss barriers between minima and their connectivity along low loss paths, \citep{2018arXiv180210026G}; \citep{2018arXiv180300885D}, to name a few examples.

One hypothesis that has gained popularity is that the flatness of minima of the loss function found by stochastic gradient-based methods results in good generalization, \citep{Hochreiter:1997:FM:258006.258007}; \citep{DBLP:journals/corr/KeskarMNST16}. \citep{2018arXiv180208770X} and \citep{DBLP:journals/corr/abs-1712-09913} measure the flatness by the spectral norm of the hessian of the loss, with respect to the parameters, at a given point in the parameter space. Both \citep{DBLP:journals/corr/abs-1710-06451} and \citep{DBLP:journals/corr/abs-1711-04623} consider the determinant of the hessian of the loss, with respect to the parameters, for the measure of flatness. For all of the work on flatness of minima cited above, authors have found that flatter minima correlate with better generalization.

In contrast to previous work on understanding the objective landscapes of neural networks in the classical supervised learning paradigm, in our work, we explore the properties of objective landscapes in the setting of gradient-based meta-learning.

\section{Gradient-based meta-learning}\label{sec:gradient_based_meta-learning}

We consider the meta-learning scenario where we have a distribution over tasks $p(\mathcal{T})$, and a model $f$ parametrized by $\theta$, that must learn to adapt to tasks $\mathcal{T}_i$ sampled from $p(\mathcal{T})$. The model is trained on a set of training tasks $\{ \mathcal{T}_i \}^{train}$ and evaluated on a set of testing tasks $\{ \mathcal{T}_i \}^{test}$, all drawn from $p(\mathcal{T})$. In this work we only consider classification tasks, with $\{ \mathcal{T}_i \}^{train}$ and $\{ \mathcal{T}_i \}^{test}$  using disjoint sets of classes to constitute their tasks. Here we consider the setting of k-shot learning, that is, when $f$ adapts to a task $\mathcal{T}_i^{test}$, it only has access to a set of few support samples $\mathcal{D}_i = \{ (\mathbf{x}_i^{(1)}, \mathbf{y}_i^{(1)}), ..., (\mathbf{x}_i^{(k)}, \mathbf{y}_i^{(k)}) \}$ drawn from $\mathcal{T}_i^{test}$. We then evaluate the model's performance on $\mathcal{T}_i^{test}$ using a new set of target samples $\mathcal{D}_i'$. By gradient-based meta-learning, we imply that $f$ is trained using information about the gradient of a certain loss function $\mathcal{L}(f(\mathcal{D}_i; \theta))$ on the tasks. Throughout this work the loss function is the cross-entropy between the predicted and true class.

\subsection{Model-Agnostic Meta-Learning (MAML)}

MAML learns an initial set of parameters $\theta$ such that on average, given a new task $\mathcal{T}_i^{test}$, only a few samples are required for $f$ to learn and generalize well to that task. During a meta-training iteration $s$, where the current parametrization of $f$ is $\theta^{s}$, a batch of $n$ training tasks is sampled from $p(\mathcal{T})$. For each task $\mathcal{T}_i$, a set of support samples $\mathcal{D}_i$ is drawn and $f$ adapts to $\mathcal{T}_i$ by performing $T$ steps of full batch gradient descent on $\mathcal{L}(f(\mathcal{D}_i; \theta))$ w.r.t. $\theta$, obtaining the adapted solution $\tilde{\theta}_i$:
\begin{equation}
\label{eq:maml:inner_loop}
    \tilde{\theta}_i = \theta^{s} - \alpha \sum_{t=0}^{T-1} \nabla_{\theta} \mathcal{L}(f(\mathcal{D}_i; \theta_i^{(t)}))
\end{equation}
where $\theta_i^{(t)} = \theta_i^{(t-1)} - \alpha \nabla_{\theta} \mathcal{L}(f(\mathcal{D}_i; \theta_i^{(t-1)}))$ and all adaptations are independent and start from $\theta^{s}$, i.e. $\theta_i^{(0)} = \theta^{s}, \forall i$. Then from each $\mathcal{T}_i$, a set of target samples $\mathcal{D}_i'$ is drawn, and the adapted meta-training solution $\theta^{s+1}$ is obtained by averaging the target gradients, such that:
\begin{equation}\label{eq:maml:outer_loop}
    \theta^{s+1} = \theta^{s} - \beta \frac{1}{n} \sum_{i=1}^{n} \nabla_{\theta} \mathcal{L}(f(\mathcal{D}_i'; \tilde{\theta}_i))
\end{equation}
As one can see in Eq.\ref{eq:maml:inner_loop} and Eq.\ref{eq:maml:outer_loop}, deriving  the meta-gradients implies computing second-order derivatives, which can come at a significant computational expense. The authors introduced a first-order approximation of MAML, where these second-order derivatives are ommited, and we refer to that other algorithm as First-Order MAML.

\subsection{Finetuning baseline}

For the finetuning baseline, the model is trained in a standard supervised learning setup: the model is trained to classify all the classes from the training split using a stochastic gradient-based optimization algorithm, its output layer size being equal to the number of meta-train classes. During evaluation on meta-test tasks, the model's final layer (fully-connected) is replaced by a layer with the appropriate size for the given meta-test task (e.g. if 5-way classification, the output layer has five logits), with its parameter values initialized to random values or with another initialization algorithm, then all the model parameters are optimized to the meta-test task, just like for the other meta-learning algorithms.

\section{Analyzing the objective landscapes}\label{sec:analyses}

\begin{figure*}[ht]
\centering
    \includegraphics[width=0.49\linewidth]{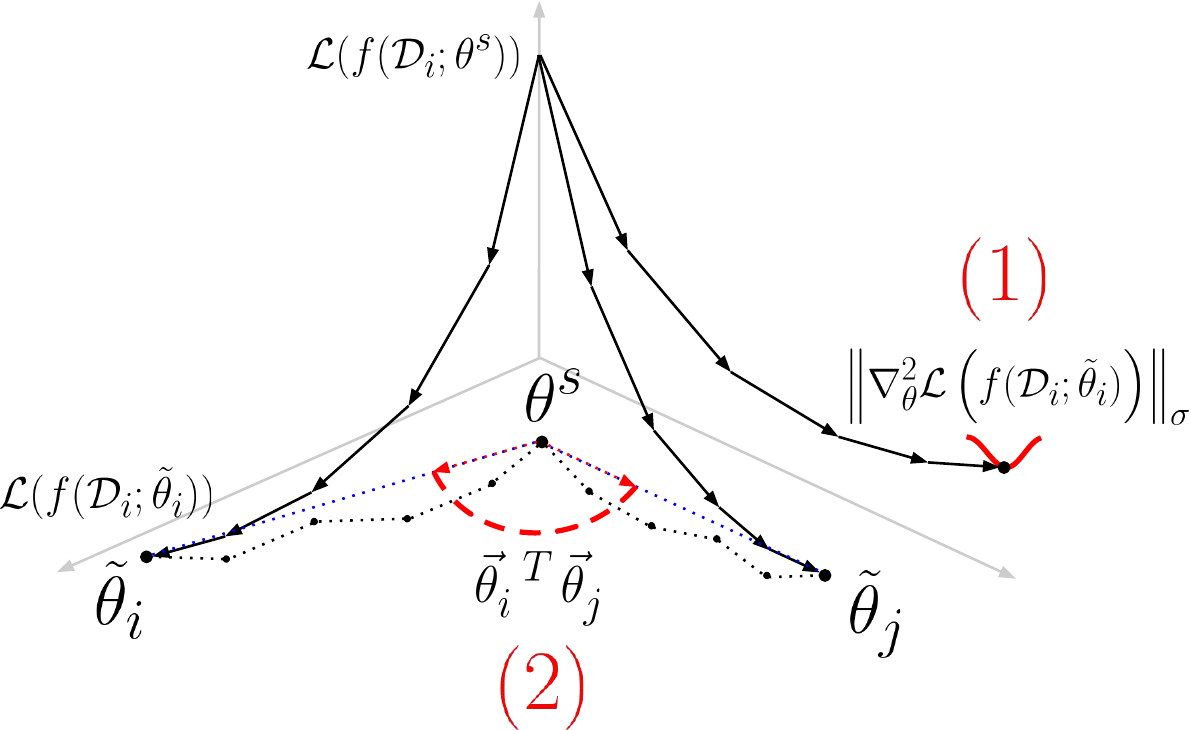}
    \includegraphics[width=0.49\linewidth]{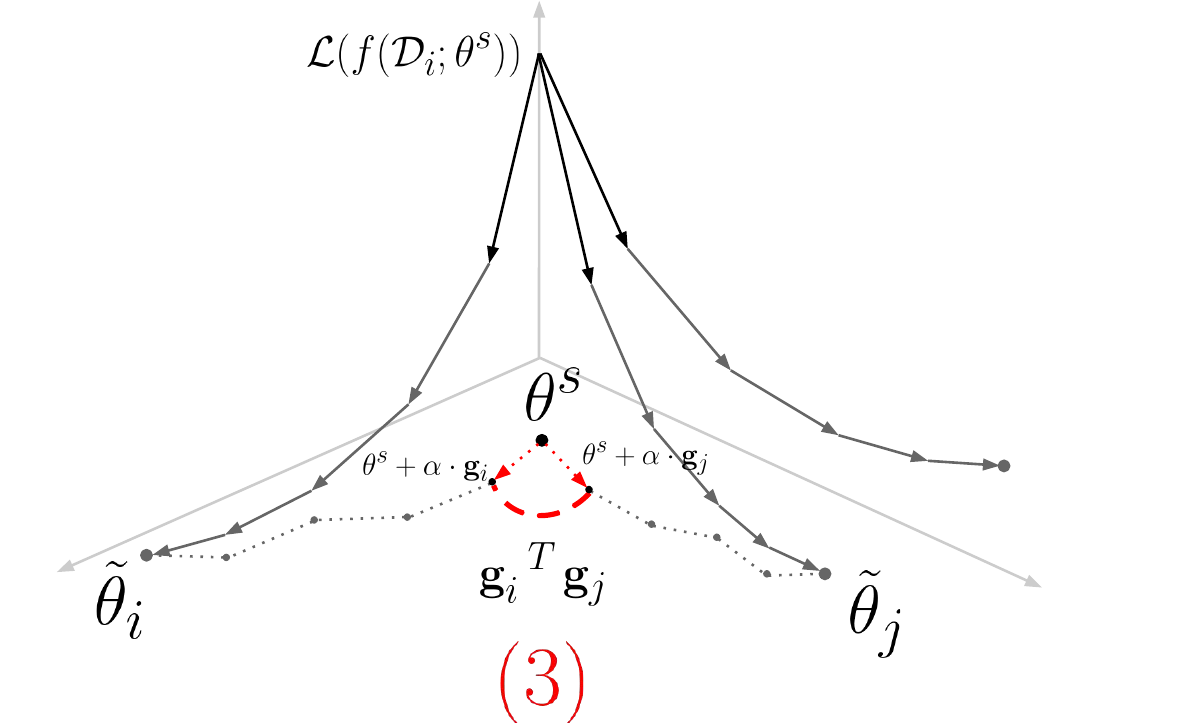}
\caption{\small Visualizations of metrics measuring properties of objective loss landscapes. The black arrows represent the descent on the support loss and the dotted lines represent the corresponding displacement in the parameter space. (1): Curvature of the loss for an adapted meta-test solution $\tilde{\theta}_i$ (for a task $\mathcal{T}_i$), is measured as the spectral norm of the hessian matrix of the loss. (2): Coherence of adaptation trajectories to different meta-test tasks is measured as the average cosine similarity for pairs of trajectory directions. A direction vector is obtained by dividing a trajectory displacement vector (from meta-train solution $\theta^s$ to meta-test solution $\tilde{\theta}_i$) by its Euclidean norm, i.e. $\vec{\theta}_i = (\tilde{\theta}_i - \theta^s) / \| \tilde{\theta}_i - \theta^s \|_2$. (3): Characterizing a meta-train solution by the coherence of the meta-test gradients, measured by the average inner product for pairs of meta-test gradient vectors $\mathbf{g}_i = -  \nabla_{\theta}\mathcal{L}(f(\mathcal{D}_i; \theta^s))$.}
\label{fig:explanatory:metrics}
\end{figure*}

In the context of gradient-based meta-learning, we define generalization as the model's ability to reach a high accuracy on a testing task $\mathcal{T}_i^{test}$, evaluated with a set of target samples $\mathcal{D}_i'$, for several testing tasks. This accuracy is computed after $f$, starting from a given meta-training parametrization $\theta^{s}$, has optimized its parameters to the task $\mathcal{T}_i^{test}$ using only a small set of support samples $\mathcal{D}_i$, resulting in the adapted solution $\tilde{\theta}_i^{test}$ (minima). We thus care about the average accuracy $\mathbb{E}_{\mathcal{T}_i^{test} \sim p(\mathcal{T})} [Acc(f(\mathcal{D}_i'; \tilde{\theta}_i^{test}) ]$. With these definitions in mind, for many meta-test tasks $\mathcal{T}_i^{test}$, we consider the optimization landscapes $\mathcal{L}(f(\mathcal{D}_i; \theta))$, and the properties of these loss landscapes evaluated at the solutions $\tilde{\theta}_i^{test}$; the adaptation trajectories when $f$, starting from $\theta^{s}$, adapts to those solutions; as well as properties of those landscapes evaluated at the meta-train solutions $\theta^s$. See Figure \ref{fig:explanatory:metrics} for a visualization of our different metrics. We follow the evolution of the metrics as meta-training progresses: after each epoch, which results in a different parametrization $\theta^s$, we adapt $f$ to several meta-test tasks, compute the metrics averaged over those tasks, and compare with $\mathbb{E} [Acc(f(\mathcal{D}_i'; \tilde{\theta}_i^{test}) ]$. We do not deal with the objective landscapes involved during meta-training, as this is beyond the scope of this work. From here on, we drop the superscript $test$ from our notation, as we exclusively deal with objective landscapes involving meta-test tasks $\mathcal{T}_i$, unless specified otherwise.

\subsection{Flatness of minima}

We start our analysis of the objective loss landscapes by measuring properties of the landscapes at the adapted meta-test solutions $\tilde{\theta}_i$. More concretely, we measure the curvature of the loss at those minima, and whether flatter minima are indicative of better generalization for the meta-test tasks.

After $s$ meta-training iterations, we have a model $f$ parametrized by $\theta^s$. During the meta-test, $f$ must adapt to several meta-test tasks $\mathcal{T}_i$ independently. For a given $\mathcal{T}_i$, $f$ adapts by performing a few steps of full-batch gradient descent on the objective landscape $\mathcal{L}(f(\mathcal{D}_i; \theta))$, using the set of support samples $\mathcal{D}_i$, and reaches an adapted solution $\tilde{\theta}_i$. Here we are interested in the curvature of $\mathcal{L}(f(\mathcal{D}_i; \tilde{\theta}_i))$, that is, the objective landscape when evaluated at such solution, and whether on average, flatter solutions favour better generalization. Considering the hessian matrix of this loss w.r.t the model parameters, defined as $H_{\theta}(\mathcal{D}_i; \tilde{\theta}_i) \doteq \nabla_{\theta}^2 \mathcal{L}(f(\mathcal{D}_i; \tilde{\theta}_i))$, we measure the curvature of the loss surface around $\tilde{\theta}_i$ using the spectral norm $ \| \cdot \|_{\sigma} $ of this hessian matrix:
\begin{equation}
    \left \| H_{\theta}(\mathcal{D}_i; \tilde{\theta}_i) \right \|_{\sigma} = \sqrt{\lambda_{max}\left( H_{\theta}(\mathcal{D}_i; \tilde{\theta}_i)^{\mathrm{H}} H_{\theta}(\mathcal{D}_i; \tilde{\theta}_i) \right)}
    =\lambda_{max} (H_{\theta}(\mathcal{D}_i; \tilde{\theta}_i))
\end{equation}
as illustrated in Figure \ref{fig:explanatory:metrics} (1). (We get $ \| H_{\theta}(\mathcal{D}_i; \tilde{\theta}_i) \|_{\sigma} = \lambda_{max} ( H_{\theta}(\mathcal{D}_i; \tilde{\theta}_i)) $ since $H_{\theta}(\mathcal{D}_i; \tilde{\theta}_i)$ is real and symmetric.)

\textit{We define the average loss curvature for meta-test solutions $\tilde{\theta}_i$, obtained from a meta-train solution $\theta^s$, as:}
\begin{equation}
    \mathbb{E}_{\mathcal{T}_i \sim p(\mathcal{T})}[ \| H_{\theta}(\mathcal{D}_i; \tilde{\theta}_i) \|_{\sigma} ]
\end{equation}

Note that we do not measure curvature of the loss at $\theta^s$, since $\theta^s$ is not a point of convergence of $f$ for the meta-test tasks. In fact, at $\theta^s$, since the model has not been adapted to the unseen meta-test classes, the target accuracy for the meta-test tasks is random chance on average. Thus, measuring the curvature of the meta-test support loss at $\theta^s$ does not relate to the notion of flatness of minima. Instead, in this work we characterize the meta-train solution $\theta^s$ by measuring the average inner product between the meta-test gradients, as explained later in Section \ref{sec:analyzing_objective_landscapes:caracterize_meta-train_solution}.

\subsection{Coherence of adaptation trajectories}\label{sec:coherence_trajectories}

Other than analyzing the objective landscapes at the different minima reached when $f$ adapts to new tasks, we also analyze the adaptation trajectories to those new tasks, and whether some similarity between them can be indicative of good generalization. 
Let's consider a model $f$ adapting to a task $\mathcal{T}_i$ by starting from $\theta^s$, moving in parameter space by performing $T$ steps of full-batch gradient descent with $\nabla_{\theta}\mathcal{L}(f(\mathcal{D}_i; \theta))$ until reaching $\tilde{\theta}_i$. We define the adaptation trajectory to a task $\mathcal{T}_i$ starting from $\theta^s$ as the sequence of iterates $(\theta^s, \theta^{(1)}_i, \theta^{(2)}_i, ..., \tilde{\theta}_i)$. To simplify the analyses and alleviate some of the challenges in dealing with trajectories of multiple steps in a parameter space of very high dimension, we define the trajectory displacement vector $(\tilde{\theta}_i - \theta^s)$. We define a trajectory direction vector $\vec{\theta}_i$ as the unit vector: $\vec{\theta}_i \doteq (\tilde{\theta}_i - \theta^s) / \| \tilde{\theta}_i - \theta^s \|_2$.

\textit{We define a metric for the coherence of adaptation trajectories to meta-test tasks $\mathcal{T}_i$, starting from a meta-train solution $\theta^s$, as the average inner product between their direction vectors:}
\begin{equation}\label{eq:trajectory_coherence}
    \mathbb{E}_{\mathcal{T}_i, \mathcal{T}_j \sim p(\mathcal{T})} [ \vec{\theta}_i^{\;\; T } \vec{\theta}_j ]
\end{equation}

The inner product between two meta-test trajectory direction vectors is illustrated in Figure \ref{fig:explanatory:metrics} (2).

\subsection{Characterizing meta-train solutions by the average inner product between meta-test gradients}\label{sec:analyzing_objective_landscapes:caracterize_meta-train_solution}

In addition to characterizing the adaptation trajectories at meta-test time, we characterize the objective landscapes at the meta-train solutions $\theta^s$. More concretely, we measure the coherence of the meta-test gradients $\nabla_{\theta}\mathcal{L}(f(\mathcal{D}_i; \theta^s))$ evaluated at $\theta^s$.

The coherence between the meta-test gradients can be viewed in relation to the metric for coherence of adaptation trajectories of Eq. \ref{eq:trajectory_coherence} from Section \ref{sec:coherence_trajectories}. Even after simplifying an adaptation trajectory by its displacement vector, measuring distances between trajectories of multiple steps in the parameter space can be problematic: because of the symmetries within the architectures of neural networks, where neurons can be permuted, different parameterizations $\theta$ can represent identically the same function $f$ that maps inputs to outputs. This problem is even more prevalent for networks with higher number of parameters. Since here we ultimately care about the functional differences that $f$ undergoes in the adaptation trajectories, measuring distances between functions in the parameter space, either using Euclidean norm or cosine similarity between direction vectors, can be problematic \citep{DBLP:journals/corr/abs-1805-08289}.

Thus to further simplify the analyses on adaptation trajectories, we can measure coherence between trajectories of only one step ($T = 1$). Since we are interested in the relation between such trajectories and the generalization performance of the models, we measure the target accuracy at those meta-test solutions obtained after only one step of gradient descent. We define those solutions as: $\theta^s + \alpha \cdot \mathbf{g}_i$, with meta-test gradient $\mathbf{g}_i = - \nabla_{\theta}\mathcal{L}(f(\mathcal{D}_i; \theta^s))$. To make meta-training consistent with meta-testing, for the meta-learning algorithms we also use $T = 1$ for the inner loop updates of Eq. \ref{eq:maml:inner_loop}.

We thus measure coherence between the meta-test gradient vectors $\mathbf{g}_i$ that lead to those solutions. Note that the learning rate $\alpha$ is constant and is the same for all experiments on a same dataset. In contrast to Section \ref{sec:coherence_trajectories}, here we observed in practice that the average inner product between meta-test gradient vectors, and not just their direction vectors, is more correlated to the average target accuracy. The resulting metric is thus the average inner product between meta-test gradients evaluated at $\theta^s$.

\textit{We define the average inner product between meta-test gradient vectors $\mathbf{g}_i$, evaluated at a meta-train solution $\theta^s$, as:}
\begin{equation}
    \mathbb{E}_{\mathcal{T}_i, \mathcal{T}_j \sim p(\mathcal{T})}[\; \mathbf{g}_i ^{\;T} \mathbf{g}_j \;]
\end{equation}

 The inner product between two meta-test gradients, evaluated at $\theta^s$, is illustrated in Figure \ref{fig:explanatory:metrics} (3). We show in the experimental results in Section  \ref{section:coherence_adaptation_trajectories} and \ref{section:coherence_meta-test_gradients} that the coherence of the adaptation trajectories, as well as of the meta-test gradients, correlate with generalization on the meta-test tasks.

\section{Experiments}\label{sec:experiments}

We apply our analyses to the two most widely used benchmark datasets for few-shot classification problems: Omniglot and MiniImagenet datasets. We use the standardized CNN architecture used by \citep{DBLP:journals/corr/VinyalsBLKW16} and \citep{DBLP:journals/corr/FinnAL17}. We perform our experiments using three different gradient-based meta-learning algorithms: MAML, First-Order MAML and a Finetuning baseline. For more details on the meta-learning datasets, architecture and meta-learning hyperparameters, see Appendix \ref{appendix:experimental_details}

We closely follow the experimental setup of \citep{DBLP:journals/corr/FinnAL17}. Except for the Finetune baseline, the meta-learning algorithms use during meta-training the same number of ways and shots as during meta-testing. For our experiments, we follow the setting of \citep{DBLP:journals/corr/VinyalsBLKW16}: for MiniImagenet, training and testing our models on 5-way classification 1-shot learning, as well as 5-way 5-shot, and for Omniglot, 5-way 1-shot; 5-way 5-shot; 20-way 1-shot; 20-way 5-shot. Each experiment was repeated for five independent runs. For the meta-learning algorithms, the choice of hyperparameters closely follows \citep{DBLP:journals/corr/FinnAL17}. For our finetuning baseline, most of the original MAML hyperparameters were left unchanged, as we want to compare the effect of the pre-training procedure, thus are kept fixed the architecture and meta-test procedures. We kept the same optimizer as for the meta-update of MAML (ADAM), and performed hyperparameter search on the mini-batch size to use, for each setting that we present. (For our reproduction results on the meta-train and meta-test accuracy, see Figure \ref{fig:accuracies:meta-train} and \ref{fig:accuracies:meta-test} in \ref{appendix:experimental_results:maml_performance}.)

\subsection{Flatness of meta-test solutions}\label{sec:experiments:flatness}

\begin{figure*}[h!]
\centering
    \subfloat[Omniglot 5-way]{%
        \includegraphics[width=0.24\linewidth]{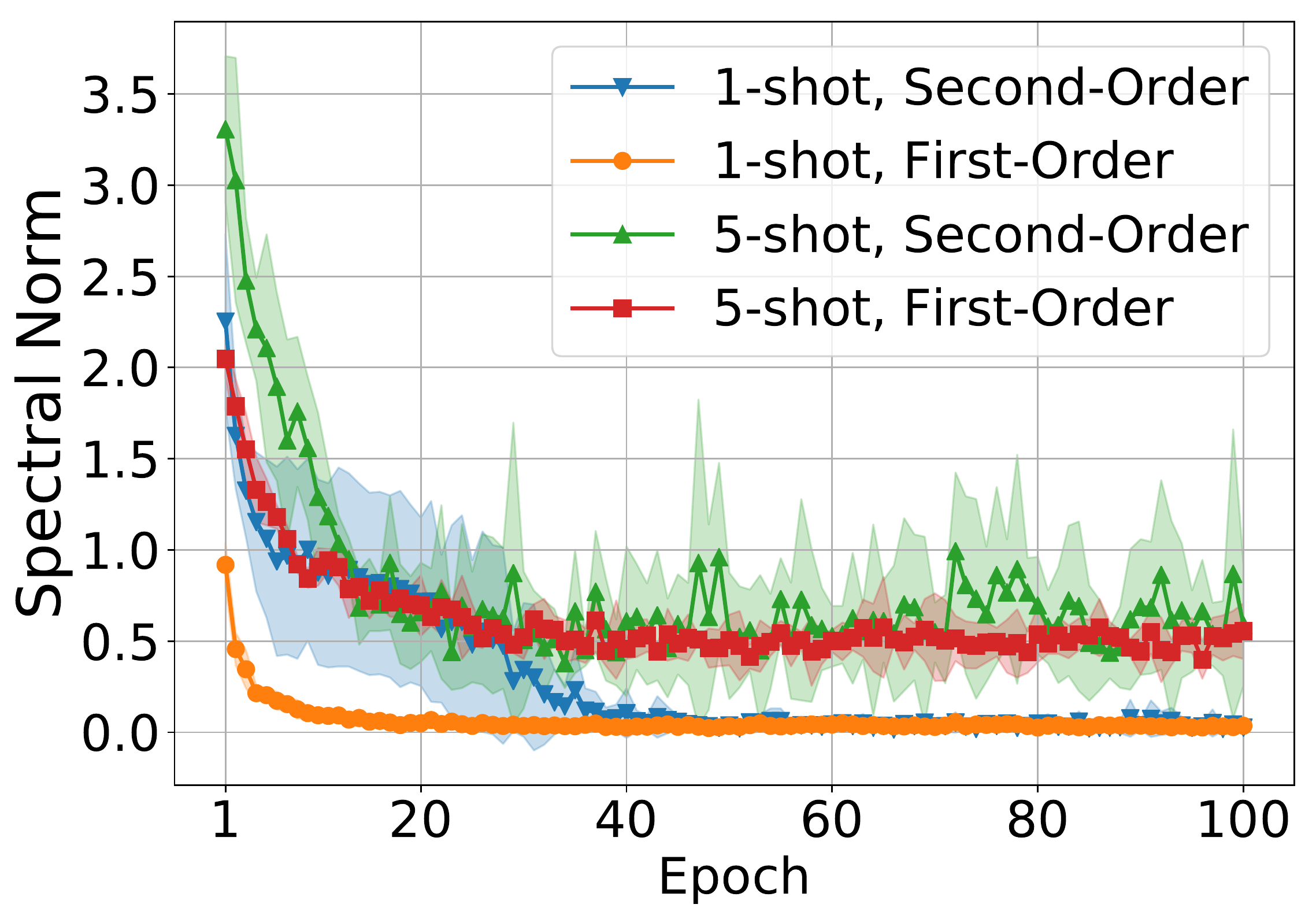}
        \label{fig:flatness:omniglot_5-way}
    }
    \subfloat[Omniglot 20-way]{%
        \includegraphics[width=0.24\linewidth]{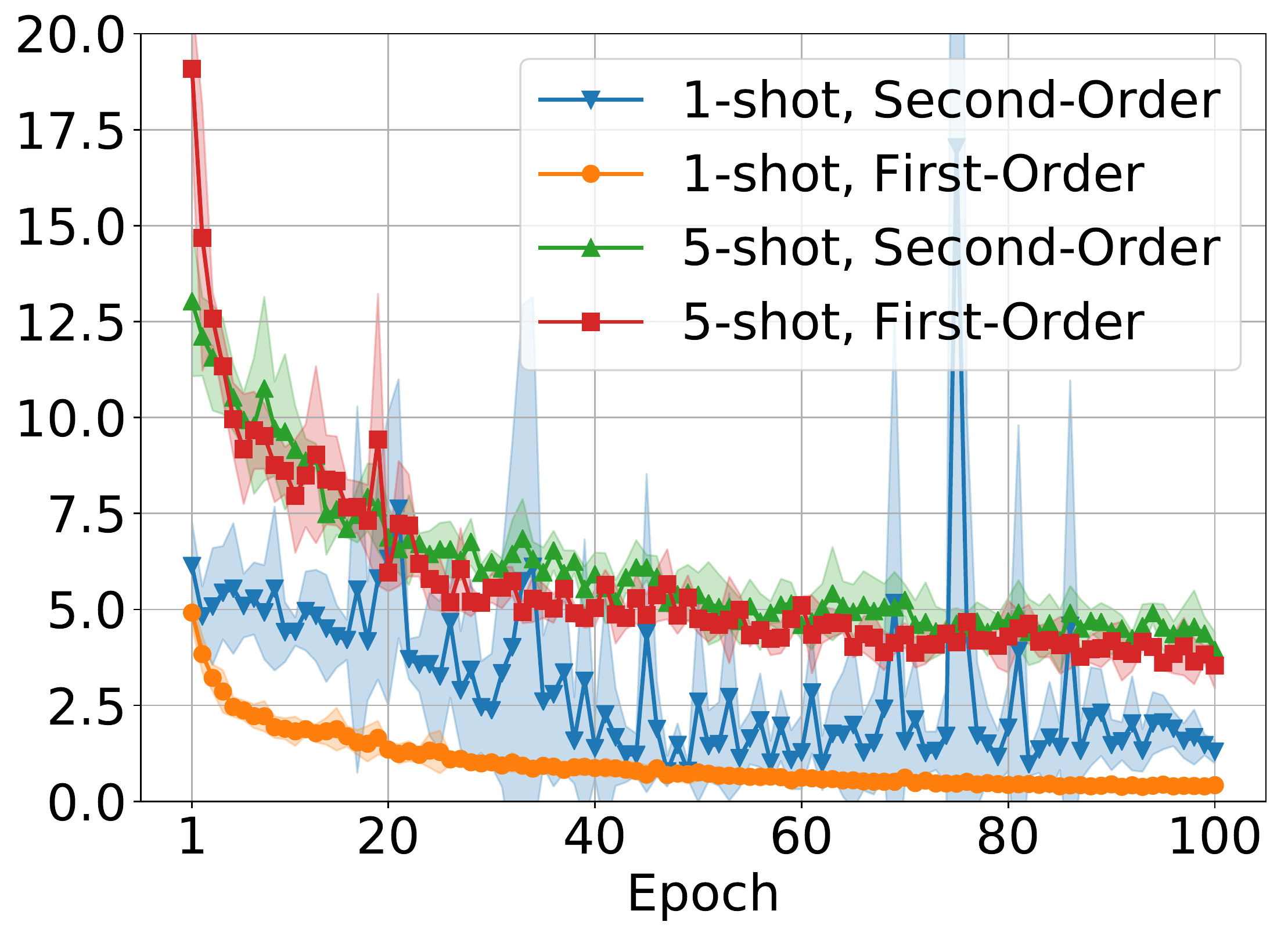}
        \label{fig:flatness:omniglot_20-way}
    }
        \subfloat[MiniImagenet \newline 5-way, 1-shot]{%
        \includegraphics[width=0.24\linewidth]{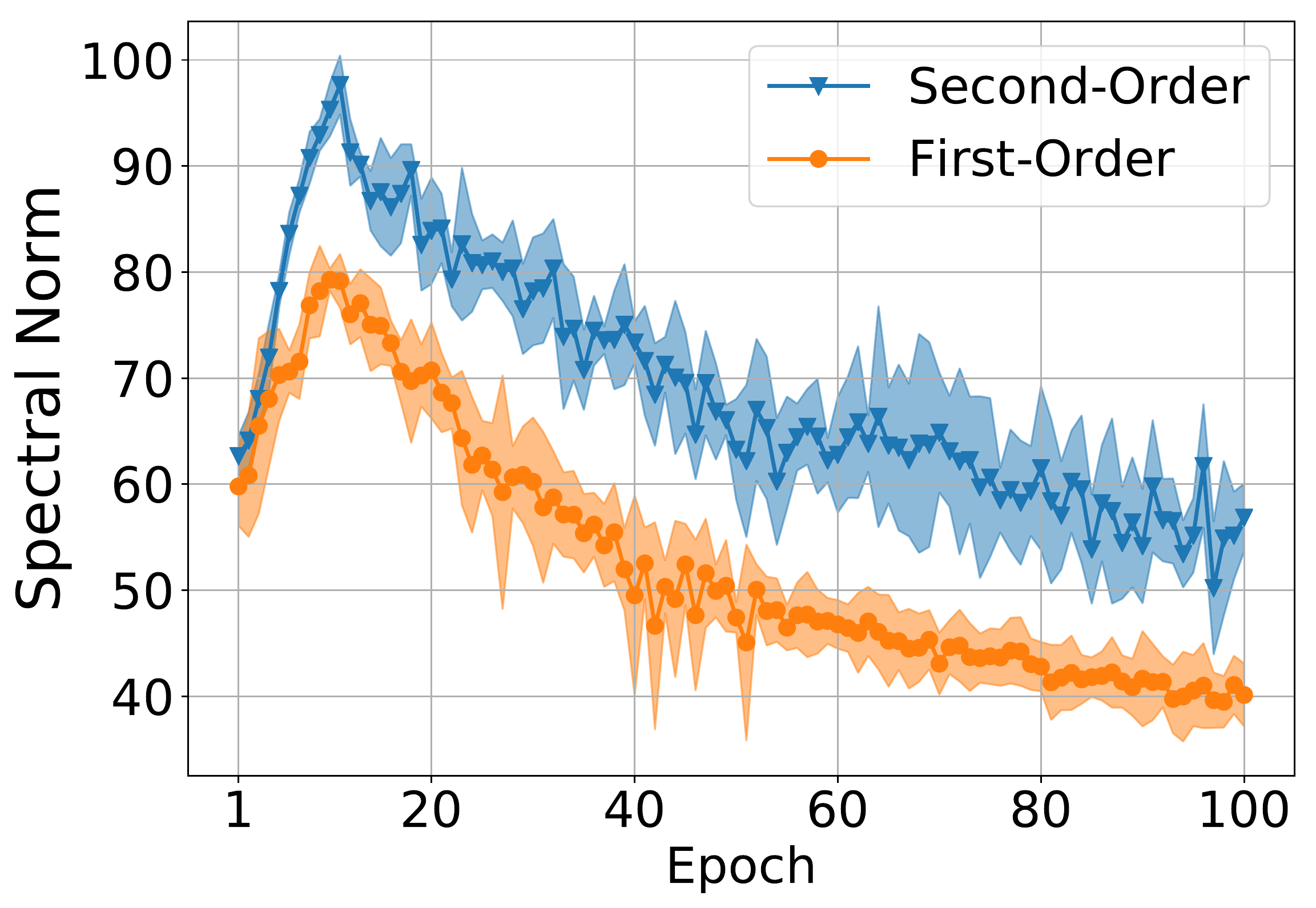}
        \label{fig:flatness:mini-imagenet_5-way_1-shot}
    }
        \subfloat[MiniImagenet \newline 5-way, 5-shot]{%
        \includegraphics[width=0.24\linewidth]{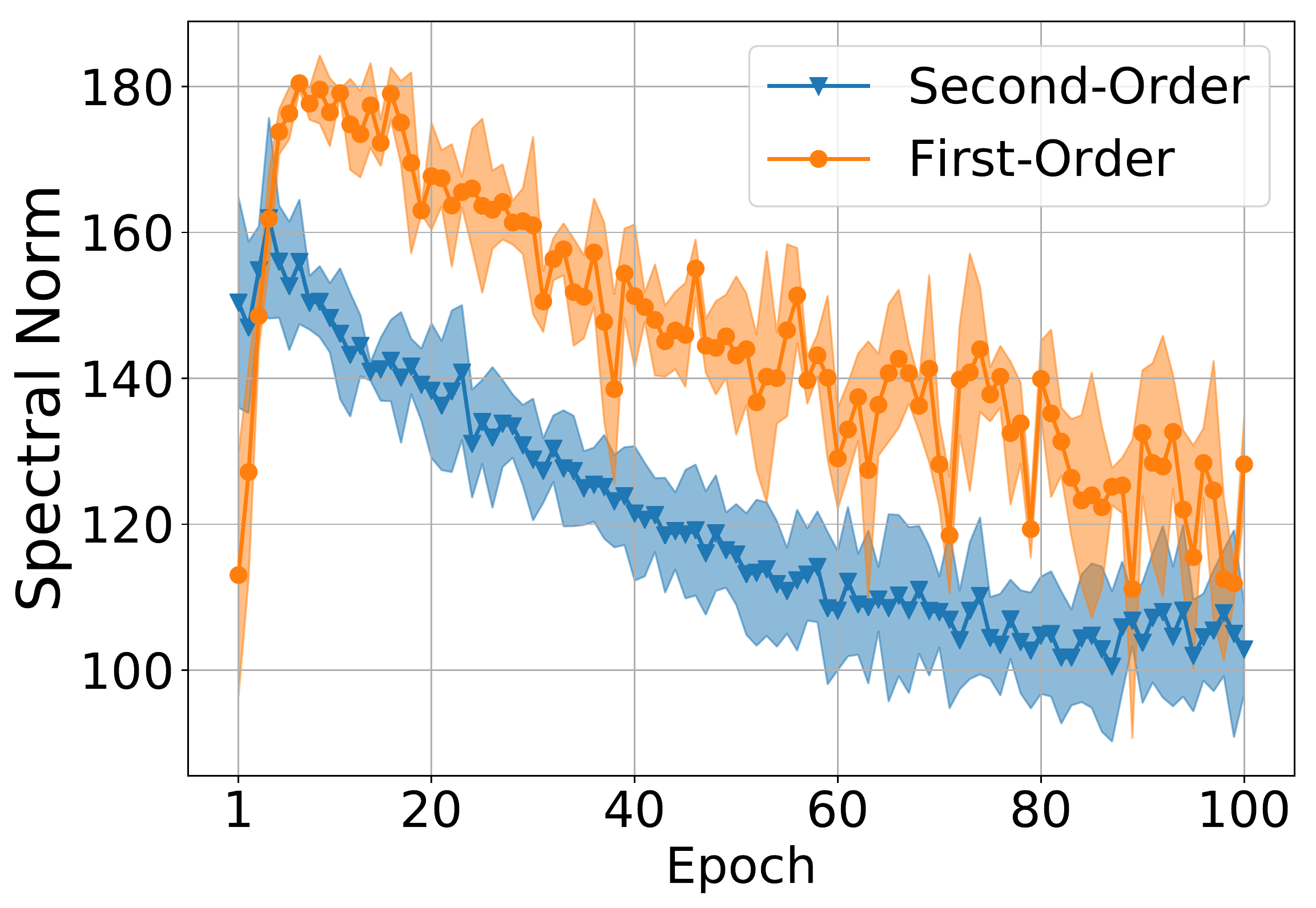}
        \label{fig:flatness:mini-imagenet_5-way_5-shot}
    }
\caption{Flatness of meta-test solutions for MAML and First-Order MAML, on Omniglot and MiniImagenet}
\label{fig:flatness_MAML}
\end{figure*}

\begin{figure}[h!]
\centering
    \subfloat[Target Accuracy]{%
        \includegraphics[width=0.3425\linewidth]{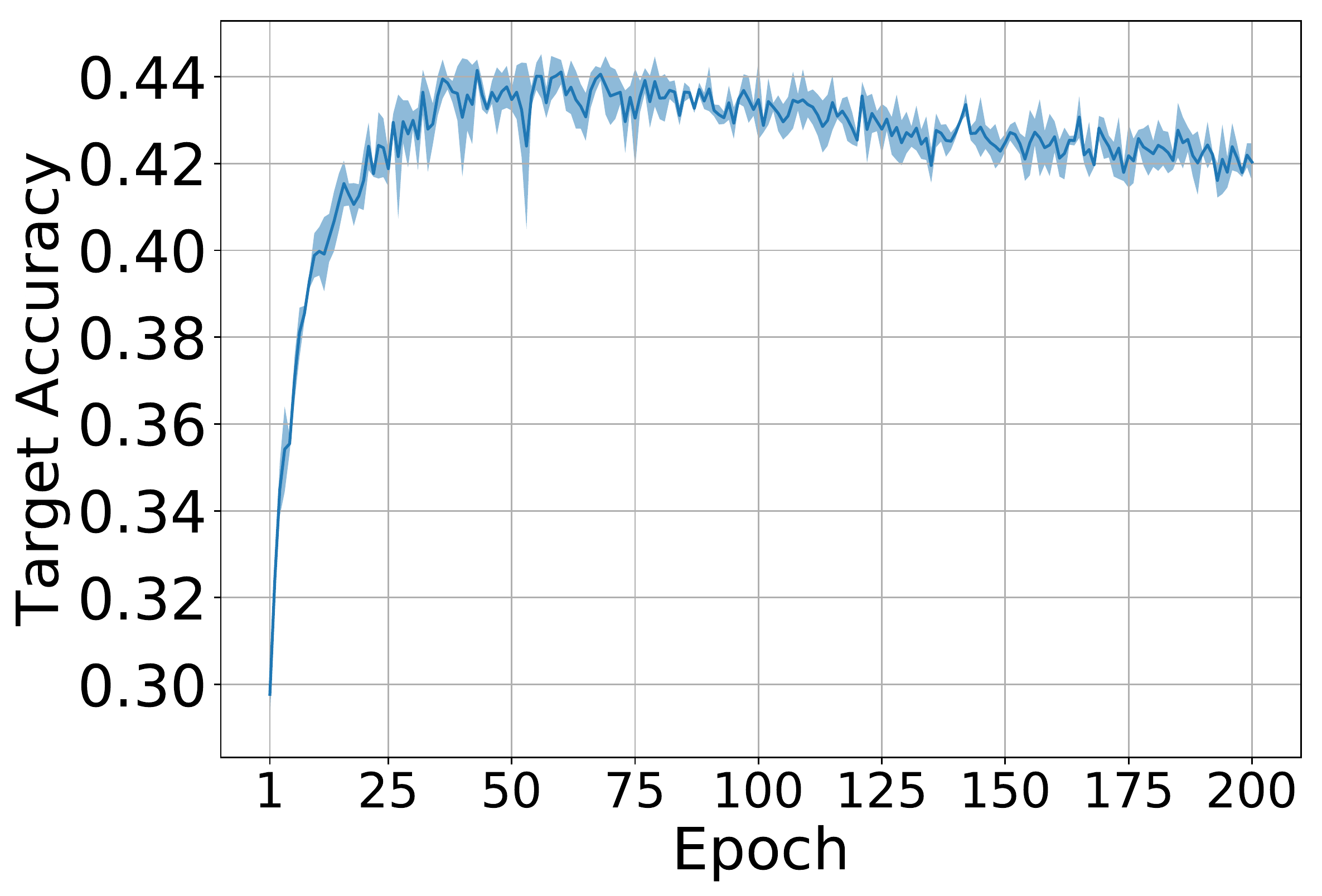}
        \label{fig:flatness_vs_gen:maml:target_acc}
    }
        \subfloat[Support loss]{%
        \includegraphics[width=0.3325\linewidth]{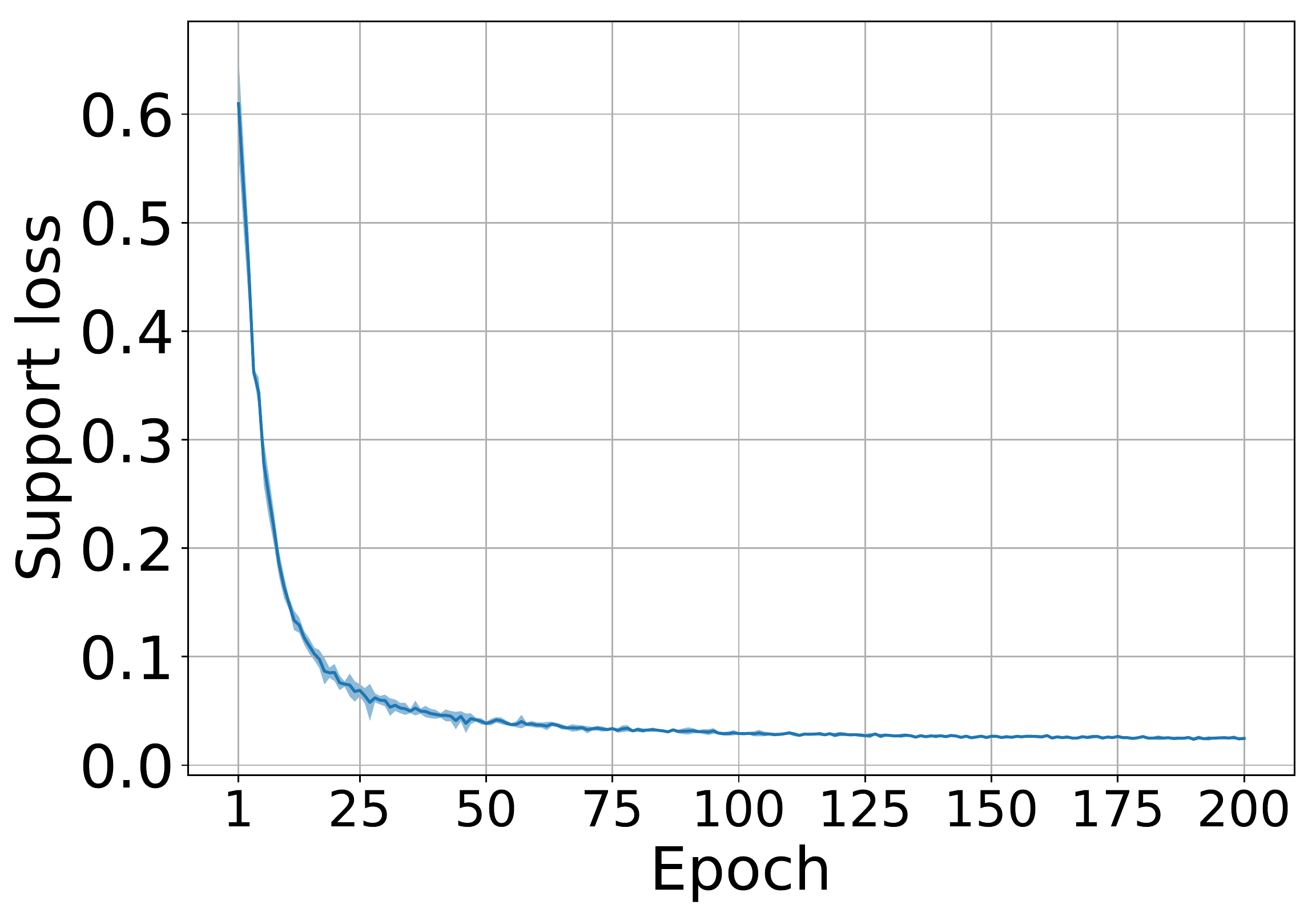}
        \label{fig:flatness_vs_gen:maml:support_loss}
    }
        \subfloat[Curvature of solutions]{%
        \includegraphics[width=0.33\linewidth]{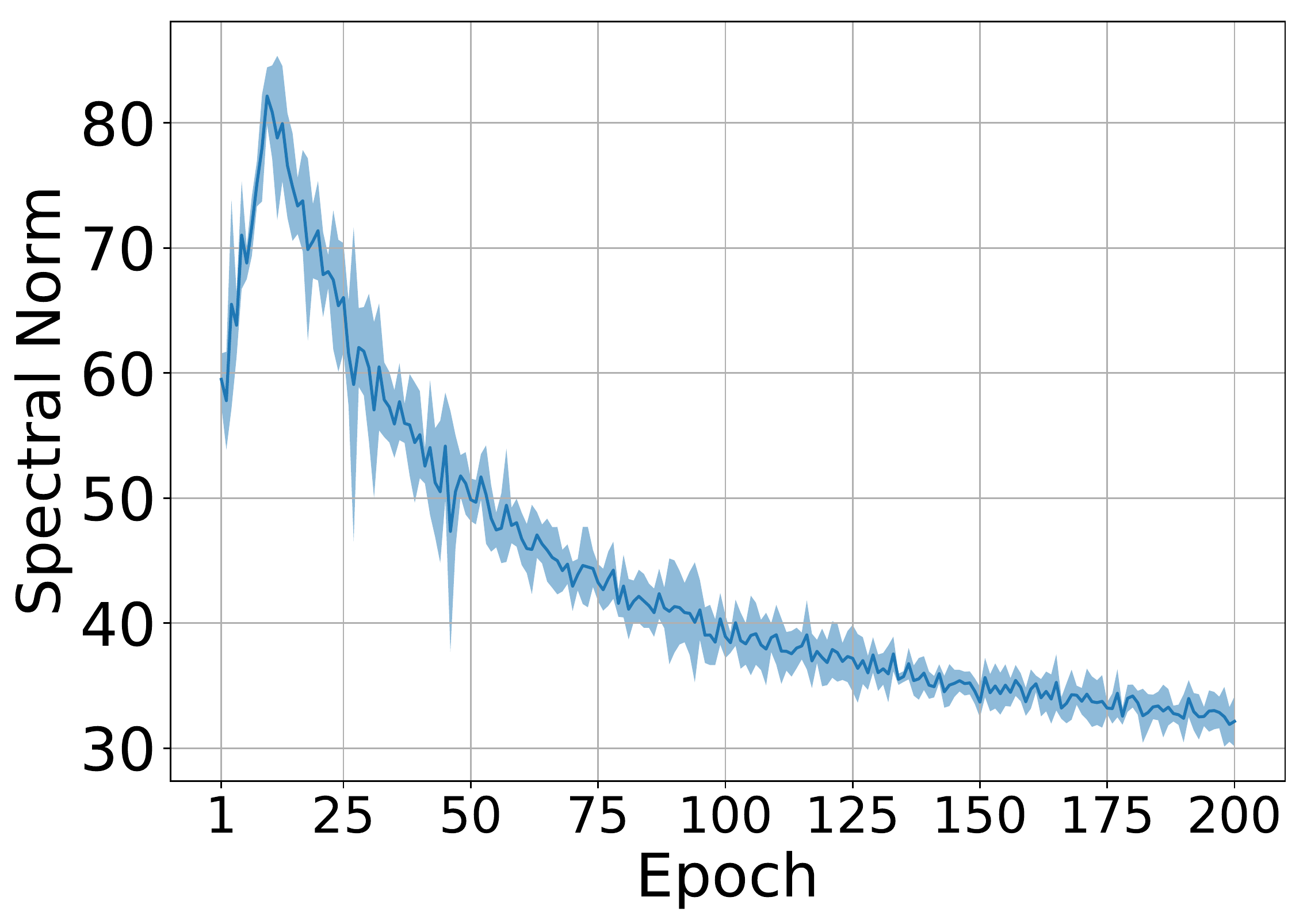}
        \label{fig:flatness_vs_gen:maml:flatness}
    }
\caption{MAML: Characterization of meta-test solutions}
\vskip -0.2in
\label{fig:flatness_vs_gen:maml}
\end{figure}

\begin{figure}[h!]
\centering
    \subfloat[Target accuracy]{%
        \includegraphics[width=0.3375\linewidth]{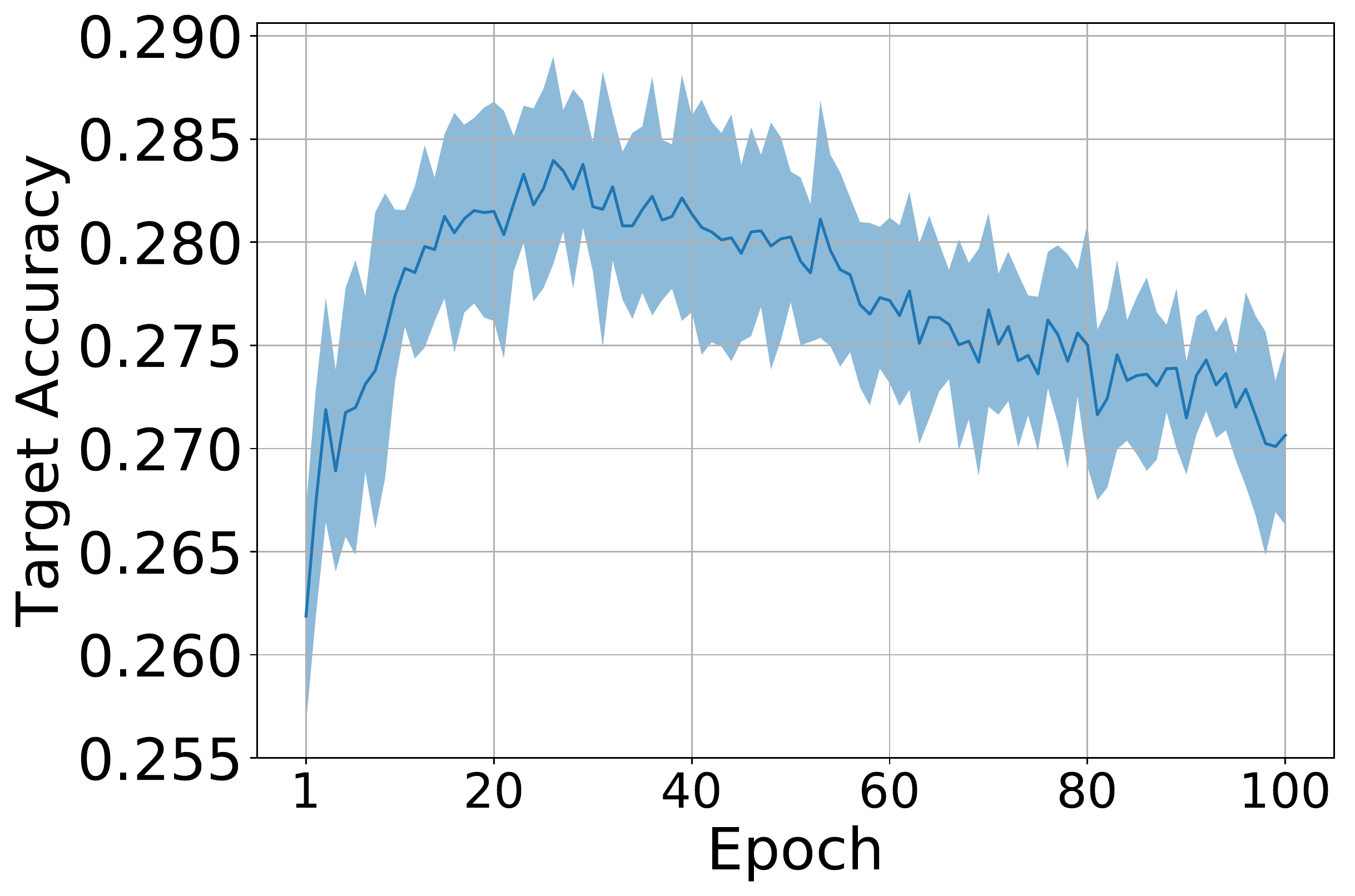}
        \label{fig:flatness_vs_gen:baseline:target_acc}
    }
        \subfloat[Support loss]{%
        \includegraphics[width=0.33\linewidth]{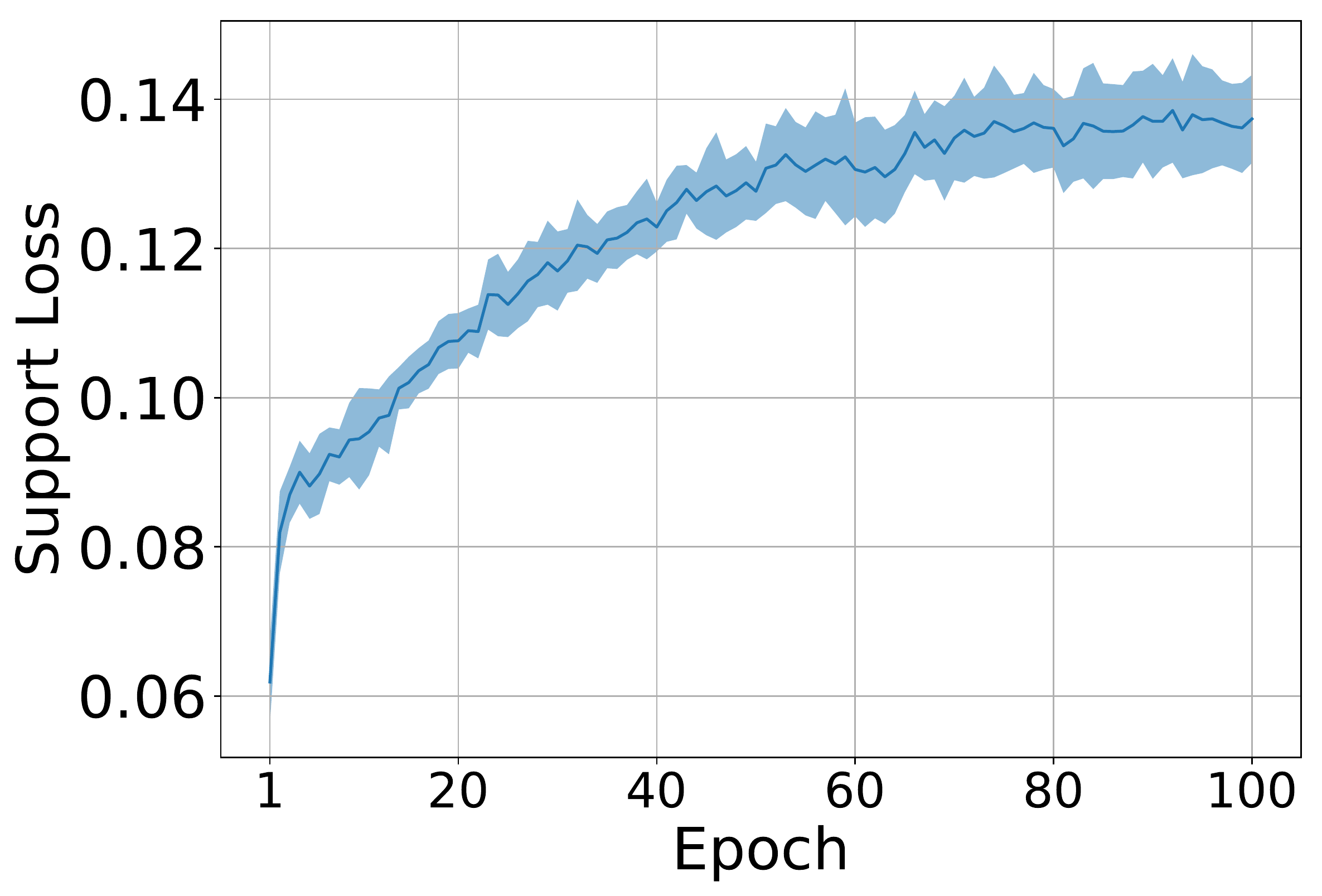}
        \label{fig:flatness_vs_gen:baseline:support_loss}
    }
        \subfloat[Curvature of solutions]{%
        \includegraphics[width=0.32\linewidth]{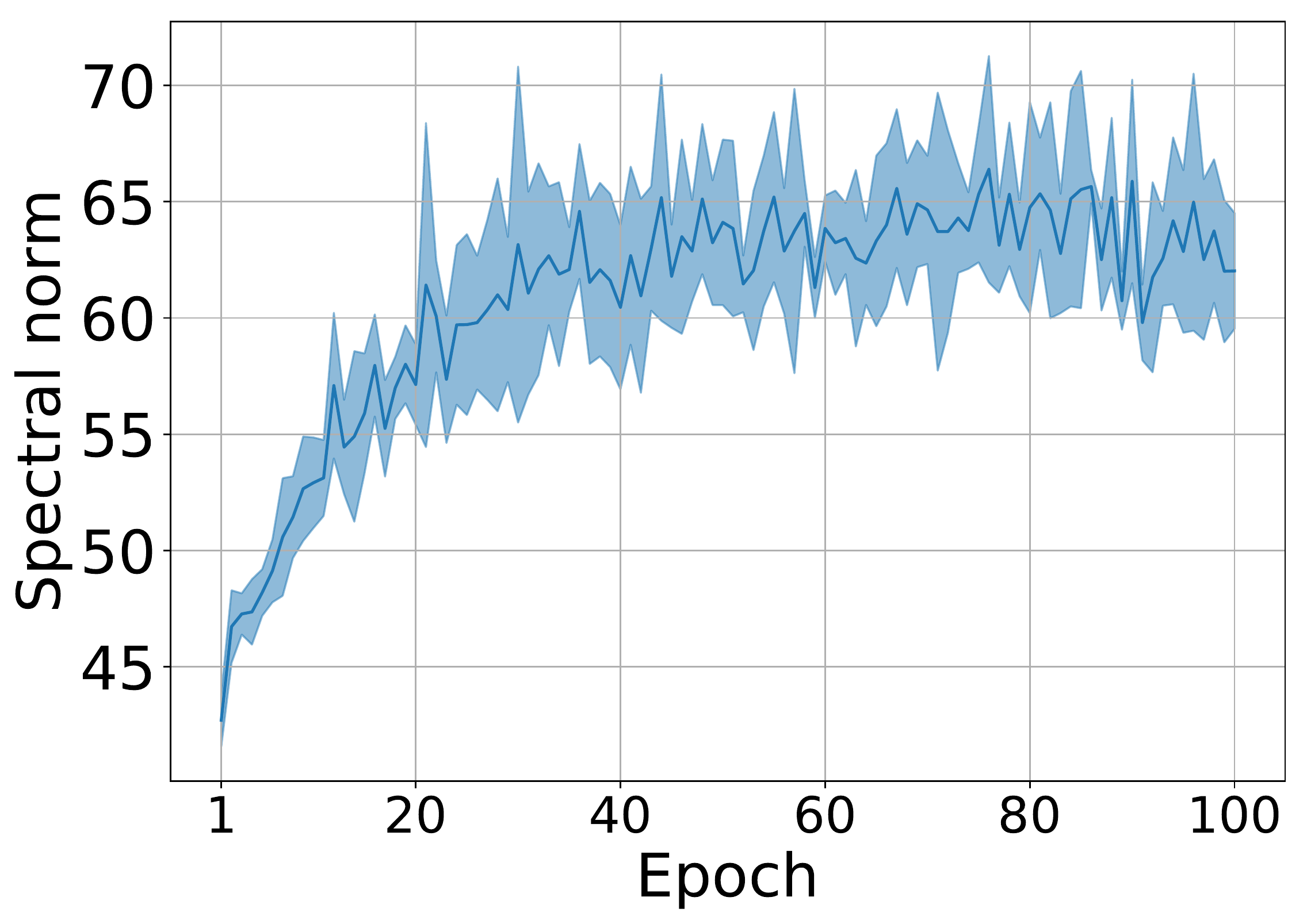}
        \label{fig:flatness_vs_gen:baseline:flatness}
    }
\caption{Finetune baseline : Characterization of meta-test solutions}
\label{fig:flatness_vs_gen:baseline}
\end{figure}

After each training epoch, we compute $\mathbb{E}[ \| H_{\theta}(\mathcal{D}_i; \tilde{\theta}_i) \|_{\sigma} ]$ using a fixed set of 60 randomly sampled meta-test tasks $\mathcal{T}_i$. Across all settings, we observe that MAML first finds sharper solutions $\tilde{\theta}_i$ until reaching a peak, then as the number of epoch grows, those solutions become flatter, as seen in Figure \ref{fig:flatness_MAML}.
To verify the correlation between $\mathbb{E}[ \| H_{\theta}(\mathcal{D}_i; \tilde{\theta}_i) \|_{\sigma} ]$ and $\mathbb{E}[ Acc(f(\mathcal{D}_i'; \tilde{\theta}_i)) ]$, we trained for an extra 100 epochs, the model that appeared the most like to overfit in a noticeable way, that is, First-Order MAML, with 5-way 1-shot learning on MiniImagenet, hoping that its decrease in $\mathbb{E}[ Acc(f(\mathcal{D}_i'; \tilde{\theta}_i)) ]$ would be reflected by an increase in $\mathbb{E}[ \| H_{\theta}(\mathcal{D}_i; \tilde{\theta}_i) \|_{\sigma} ]$ after a certain point. On the contrary, and remarkably, even as $f$ starts to show poorer generalization (see Figure \ref{fig:flatness_vs_gen:maml:target_acc}), the solutions keep getting flatter, as shown in Figure \ref{fig:flatness_vs_gen:maml:flatness}. Thus for the case of gradient-based meta-learning, our finding directly contradicts the argument that flatter minima favour better generalization.
We performed the same analysis for our finetuning baseline (Figures \ref{fig:flatness_vs_gen:baseline:target_acc}, \ref{fig:flatness_vs_gen:baseline:flatness}), with results suggesting that flatness of solutions might be more linked with $\mathbb{E}[ \mathcal{L}(f(\mathcal{D}_i; \tilde{\theta}_i)) ]$, the average level of support loss attained by the solutions $\tilde{\theta}_i$ (see Figures \ref{fig:flatness_vs_gen:baseline:support_loss} and \ref{fig:flatness_vs_gen:maml:support_loss}), which is not an indicator for generalization. We also noted that across all settings involving MAML and First-Order MAML, this average meta-test support loss $\mathbb{E}[ \mathcal{L}(f(\mathcal{D}_i; \tilde{\theta}_i)) ]$ decreases monotonically as meta-training progresses.

\newpage

\subsection{Coherence of adaptation trajectories}\label{section:coherence_adaptation_trajectories}

\begin{figure}[h!]
\centering
    \subfloat[MiniImagenet, 5-way, 1-shot, First-Order]{%
        \includegraphics[width=0.49\linewidth]{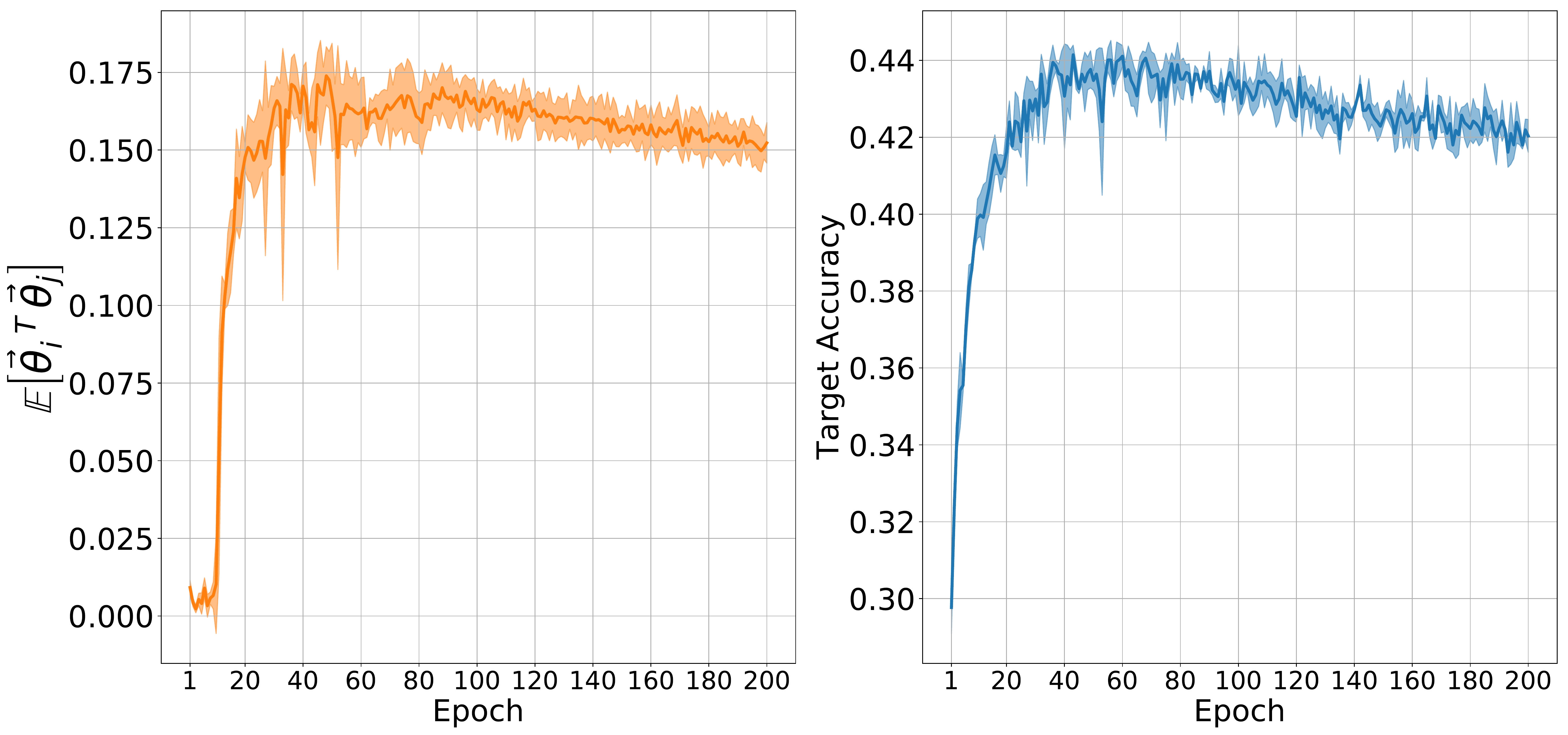}
        \label{fig:angles_vs_gen:maml:mini_5w1s_first-order}
    }
    \subfloat[MiniImagenet, 5-way, 1-shot, Second-Order]{%
        \includegraphics[width=0.49\linewidth]{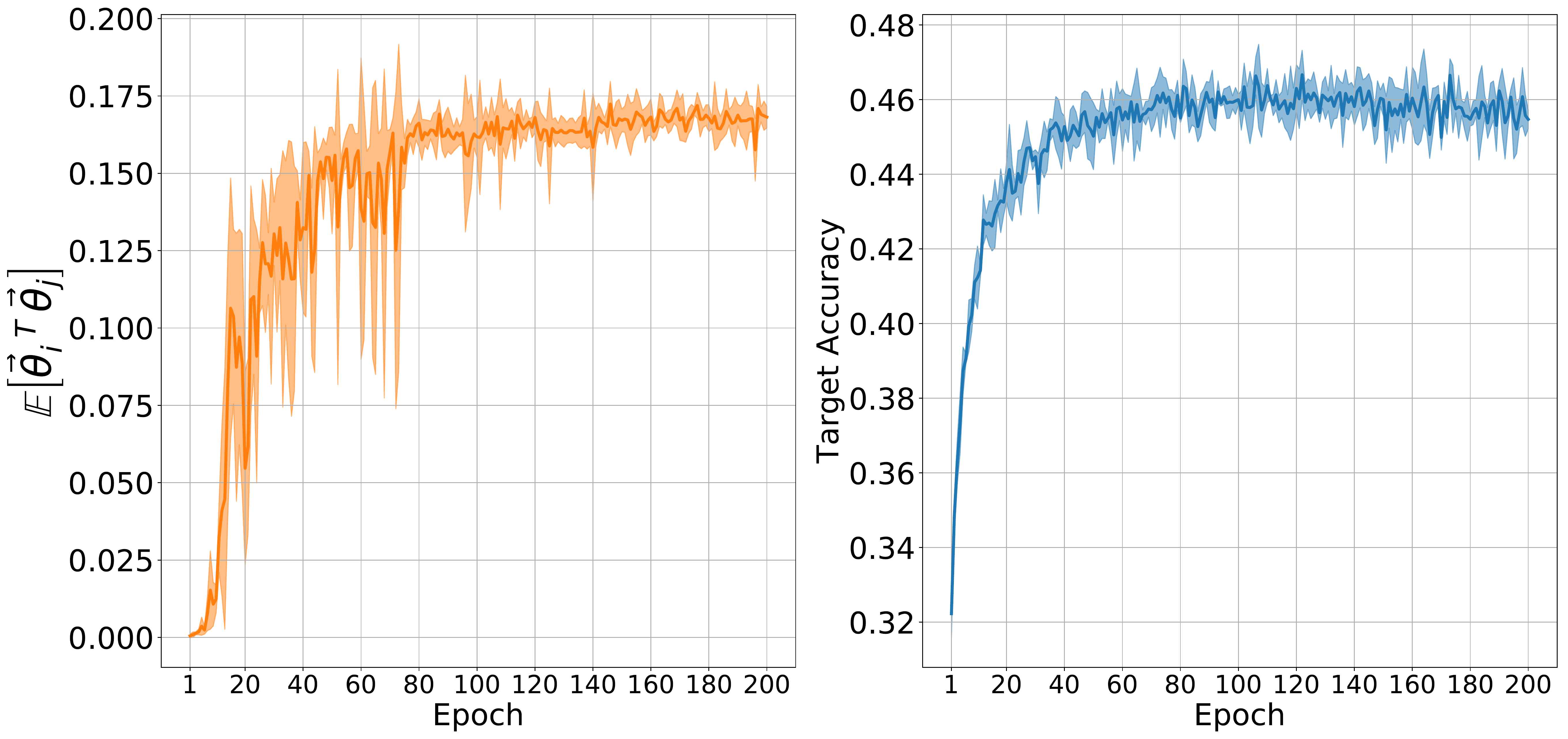}
        \label{fig:angles_vs_gen:maml:mini_5w1s}
    }
\caption{Comparison between average inner product between meta-test trajectory direction vectors (orange), and average target accuracy on meta-test tasks (blue), MAML First-Order and Second-Order, MiniImagenet 5-way 1-shot. See Figure \ref{fig:appen:results:angles_vs_gen} in Appendix \ref{appen:angle} for full set of experiments.}
\label{fig:angles_vs_gen}
\end{figure}

In this section, we use the same experimental setup as in Section \ref{sec:experiments:flatness}, except here we measure $\mathbb{E} [ \vec{\theta}_i^{\;\; T } \vec{\theta}_j ]$. To reduce the variance on our results, we sample 500 tasks after each meta-training epoch. Also for experiments on Omniglot, we drop the analyses with First-Order MAML, since it yields performance very similar to that of the Second-Order MAML. We start our analyses with the setting of "MiniImagenet, First-Order MAML, 5-way 1-shot", as it allowed us to test and invalidate the correlation between flatness of solutions and generalization, earlier in Section \ref{sec:experiments:flatness}.

\begin{figure}[h!]
\centering
    \subfloat[Trajectories coherence]{%
        \includegraphics[width=0.23\linewidth]{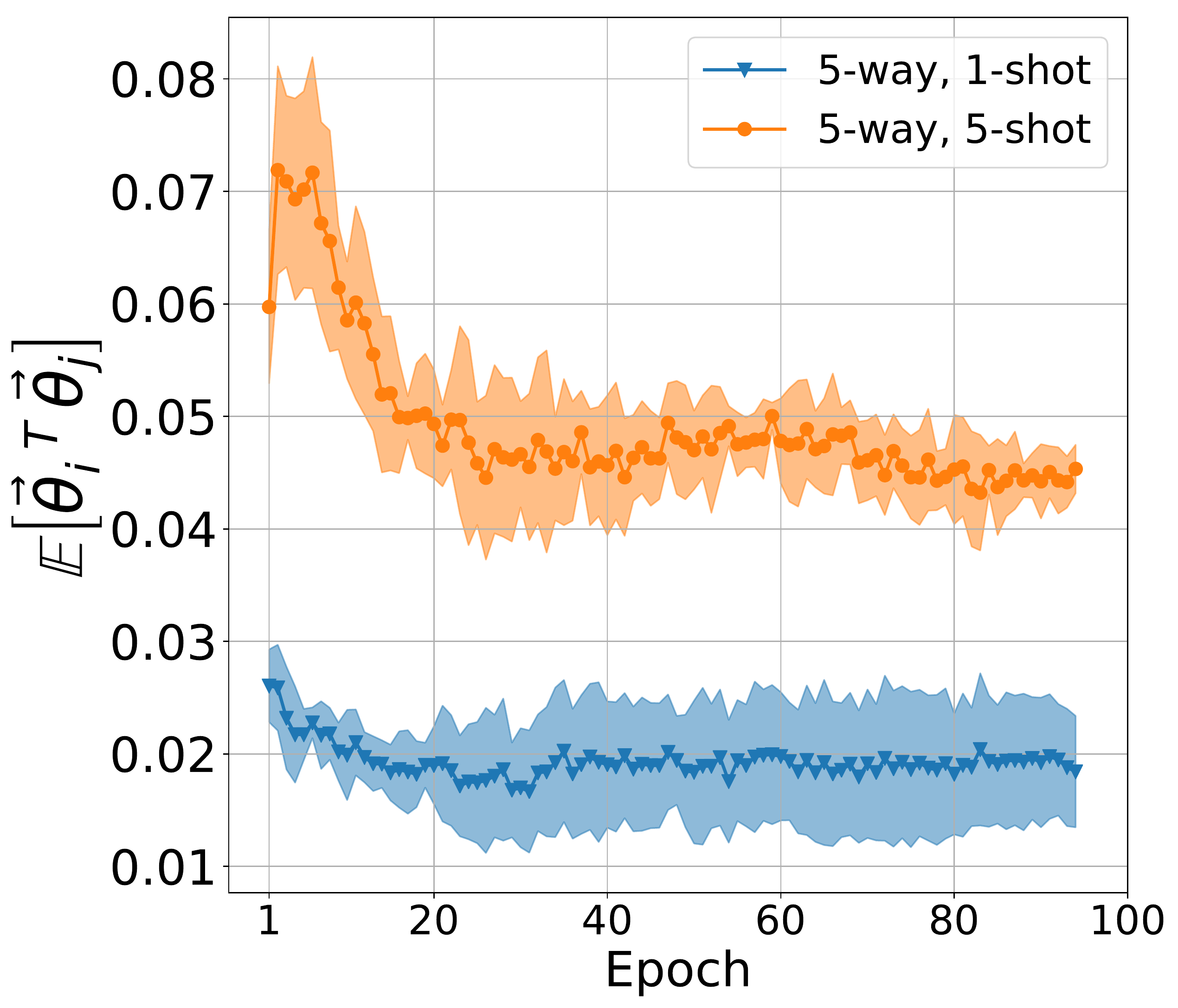}
        \label{fig:divergence:baseline:angles}
    }
    \subfloat[Gradients coherence]{%
        \includegraphics[width=0.24\linewidth]{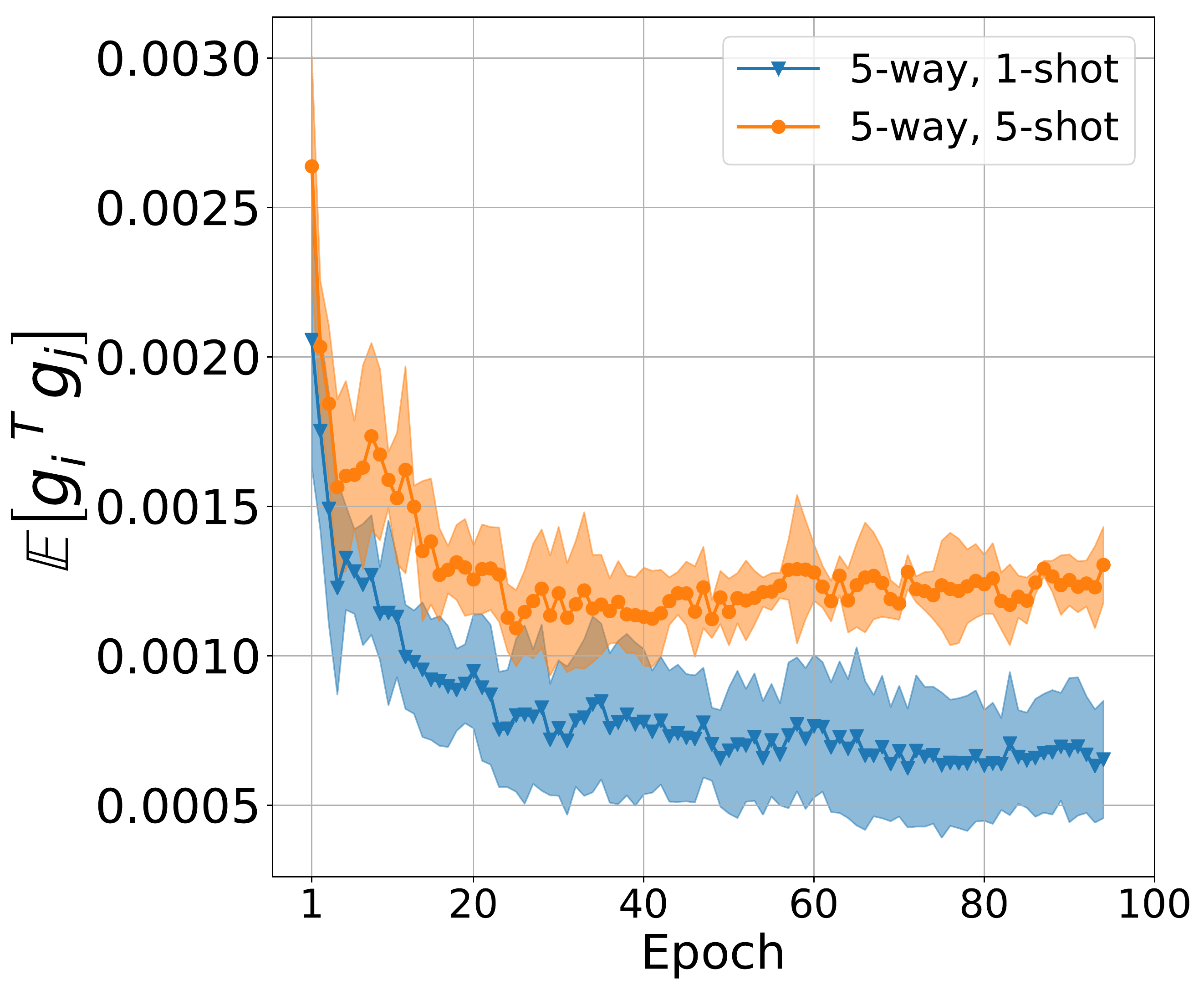}
        \label{fig:divergence:baseline:inner_products}
    }
    \subfloat[$l_2$ norm of trajectories (1-shot)]{%
        \includegraphics[width=0.23\linewidth]{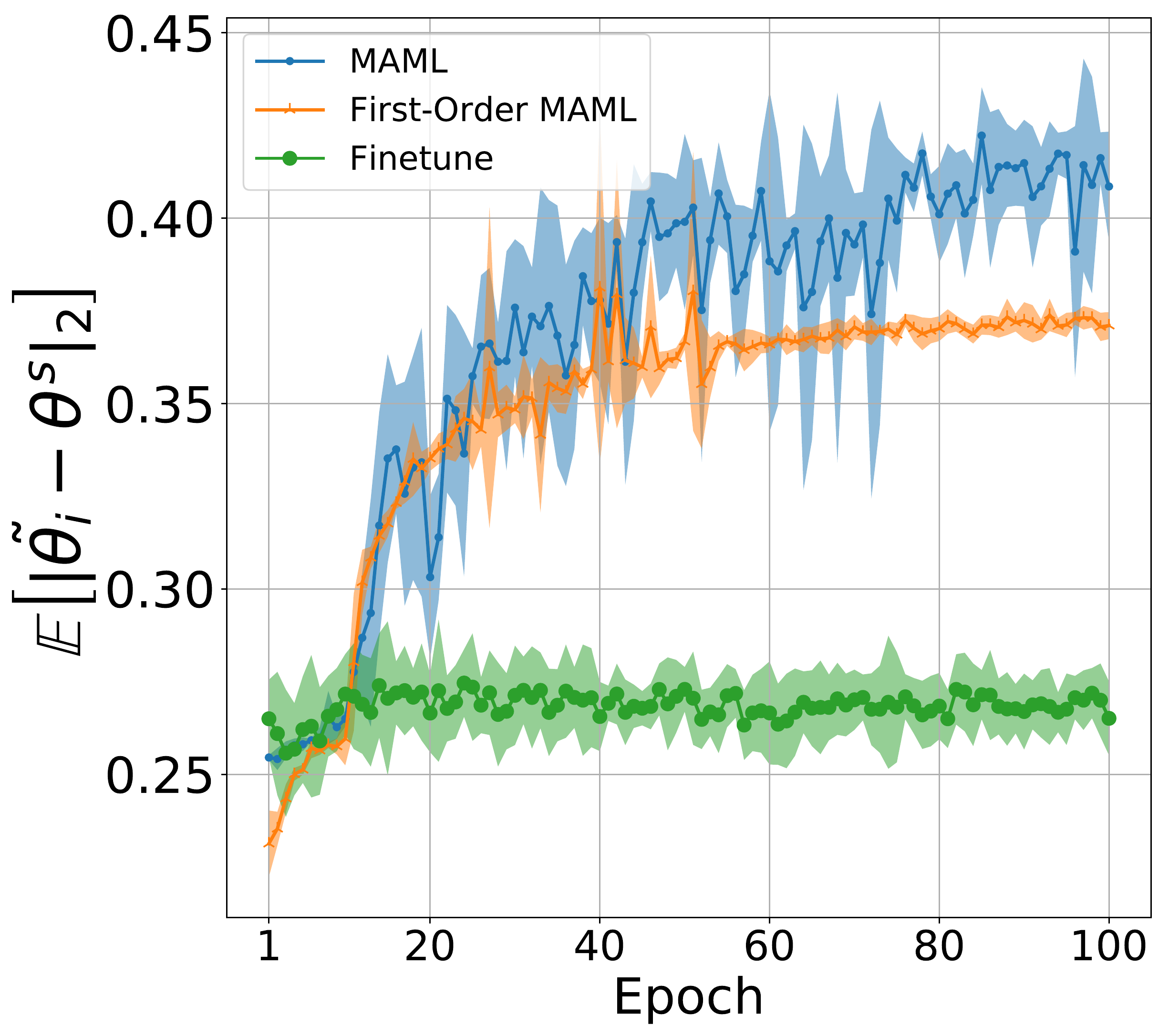}
        \label{fig:l2norm:1-shot}
    }
    \subfloat[$l_2$ norm of trajectories (5-shot)]{%
        \includegraphics[width=0.225\linewidth]{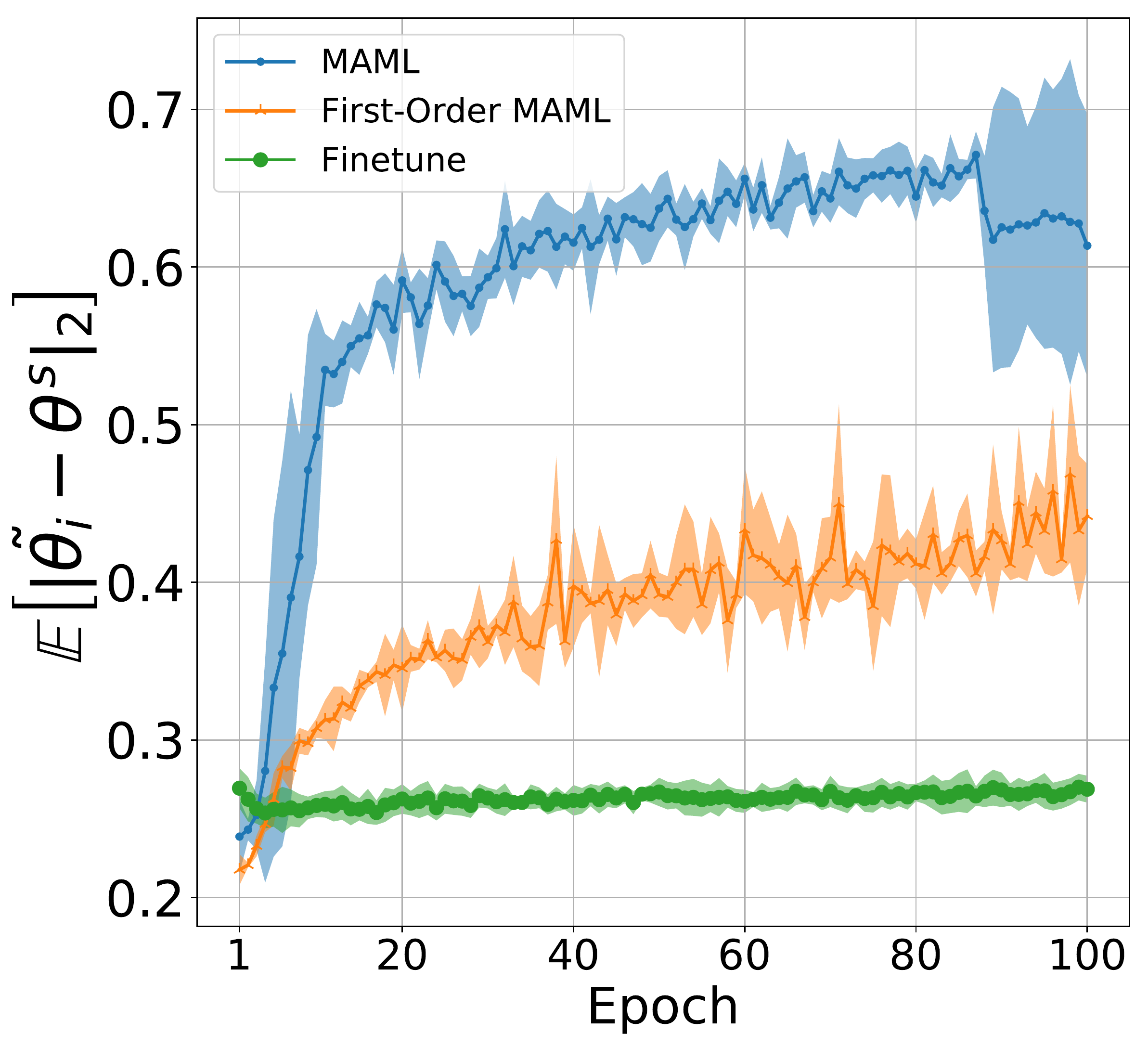}
        \label{fig:l2norm:5-shot}
    }
\caption{\protect\subref{fig:divergence:baseline:angles}: Average inner product between meta-test adaptation direction vectors, for Finetuning baseline on MiniImagenet. \protect\subref{fig:divergence:baseline:inner_products}: Average inner product between meta-test gradients, for Finetuning baseline on MiniImagenet. Average $l_2$ norm of meta-test adaptation trajectories, all algorithms on MiniImagenet, \protect\subref{fig:l2norm:1-shot}: 1-shot learning, \protect\subref{fig:l2norm:5-shot}: 5-shot learning.}
\label{fig:divergence:baseline}
\end{figure}

We clearly observe a correlation between the coherence of adaptation trajectories and generalization to new tasks, with higher average inner product between trajectory directions, thus smaller angles, being linked to higher average target accuracy on those new tasks, as shown in Figure \ref{fig:angles_vs_gen:maml:mini_5w1s_first-order}. We then performed the analysis on the other settings, with the same observations (see Figure \ref{fig:angles_vs_gen:maml:mini_5w1s} and Figure \ref{fig:appen:results:angles_vs_gen} in Appendix \ref{appen:angle} for full set of experiments). We also perform the analysis on the Finetuning baselines, which reach much lower target accuracies, and where we see that $\mathbb{E} [ \vec{\theta}_i^{\;\; T } \vec{\theta}_j ]$ remains much closer to zero, meaning that trajectory directions are roughly orthogonal to each other, akin to random vectors in high dimension (see Figure \ref{fig:divergence:baseline:angles}).

\subsection{Characterizing meta-train solutions by the average inner product between meta-test gradients}\label{section:coherence_meta-test_gradients}

\begin{figure*}[h!]
\centering
    \subfloat[MiniImagenet, 5-way, 5-shot, First-Order]{%
        \includegraphics[width=0.49\linewidth]{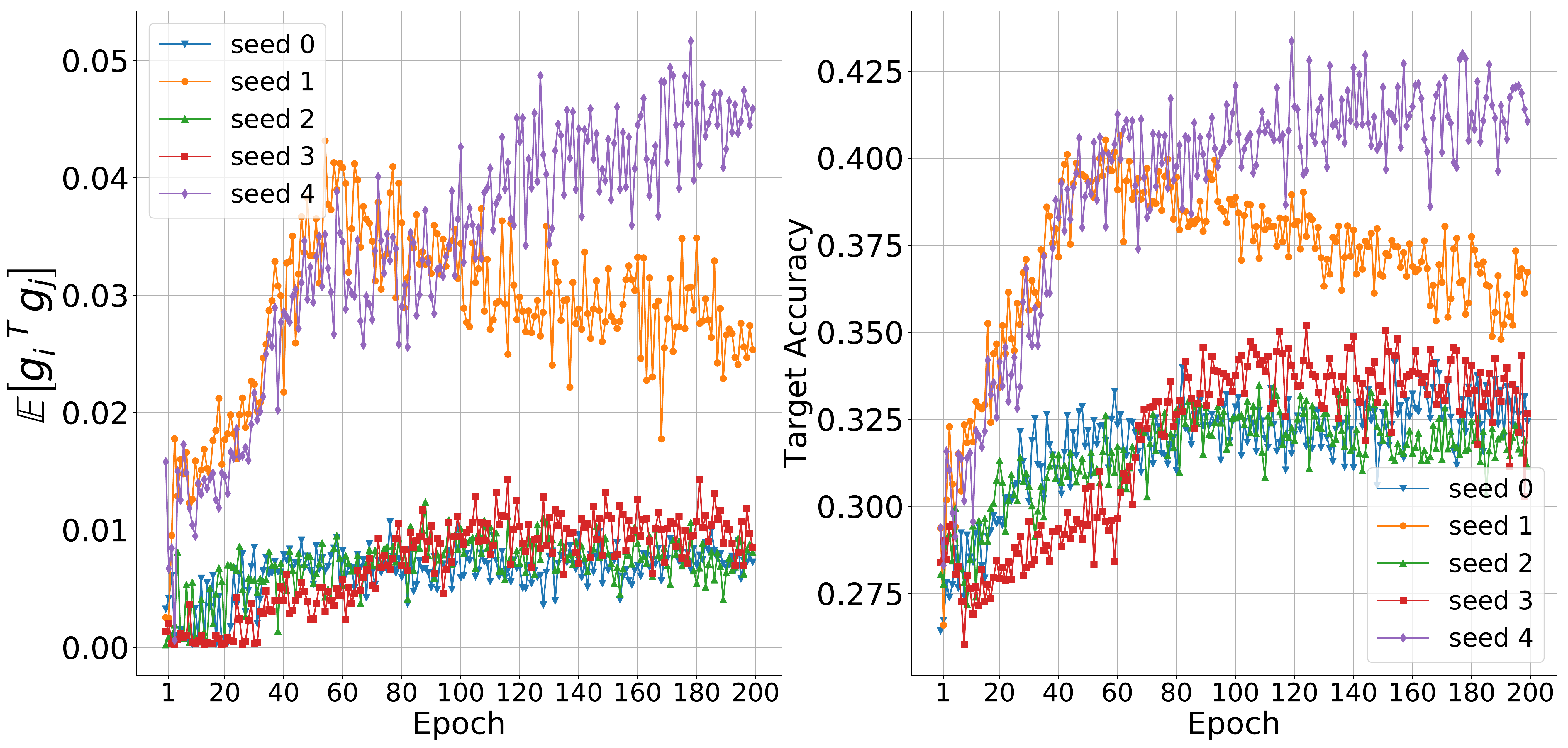}
        \label{fig:inner-products_vs_gen:mini_5w5s_first-order}
    }
    \subfloat[MiniImagenet, 5-way, 5-shot, Second-Order]{%
        \includegraphics[width=0.49\linewidth]{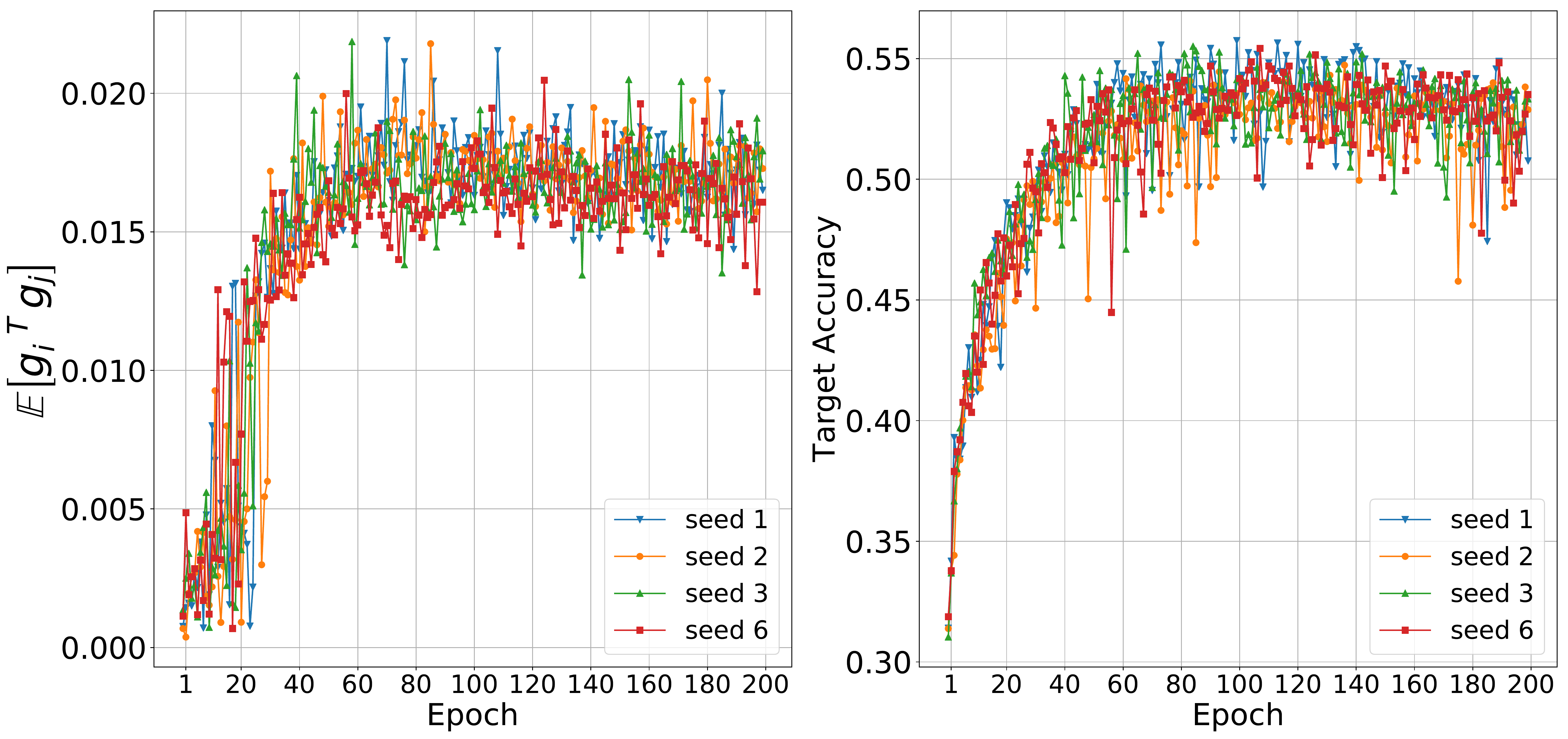}
        \label{fig:inner-products_vs_gen:mini_5w5s}
        }
\caption{Comparison between average inner product between meta-test gradient vectors, evaluated at meta-train solution, and average target accuracy on meta-test tasks, with higher average inner product being linked to better generalization. See Figure \ref{fig:appen:results:dotprods_vs_gen} in Appendix \ref{appen:scalarprod} for full set of experiments.}
\label{fig:inner-products_vs_gen}
\end{figure*}

Despite the clear correlation between $\mathbb{E} [ \vec{\theta}_i^{\;\; T } \vec{\theta}_j ]$ and generalization for the settings that we show in Figure \ref{fig:angles_vs_gen} and \ref{fig:appen:results:angles_vs_gen}, we observed that for some other settings, this relationship appears less linear. We conjecture that such behavior might arise from the difficulties of measuring distances between networks in the parameter space, as explained in Section \ref{sec:analyzing_objective_landscapes:caracterize_meta-train_solution}. Here we present our results on the characterization of the objective landscapes at the meta-train solutions $\theta^s$, by measuring the average inner product between meta-test gradient vectors $\mathbf{g}_i$.

We observe that coherence between meta-test gradients is correlated to generalization, which is consistent with the observations on the coherence of adaptation trajectories from Section \ref{section:coherence_adaptation_trajectories}. In Figure \ref{fig:inner-products_vs_gen}, we compare $\mathbb{E}[\; \mathbf{g}_i ^{\;T} \mathbf{g}_j \;]$ to the target accuracy (here we show results for individual model runs rather than the averages over the runs). See Figure \ref{fig:appen:results:dotprods_vs_gen} in Appendix \ref{appen:scalarprod} for the full set of experiments. This metric consistently correlates with generalization across the different settings. Similarly as in Section \ref{section:coherence_adaptation_trajectories}, for our finetuning baselines we observe very low coherence between meta-test gradients (see Figure \ref{fig:divergence:baseline:inner_products}). 

Based on the observations we make in Section \ref{section:coherence_adaptation_trajectories} and \ref{section:coherence_meta-test_gradients}, we propose to regularize gradient-based meta-learning as described in Section \ref{section: algo}. As an added observation, here we include our experimental results on the average meta-test trajectory norm $\mathbb{E} [ \| \tilde{\theta}_i - \theta^s \|_2 ] $ (where we used $T=5$), in Figure \ref{fig:l2norm:1-shot} and \ref{fig:l2norm:5-shot}, where $\mathbb{E} [ \| \tilde{\theta}_i - \theta^s \|_2 ] $ grows as meta-training progresses when $f$ is meta-trained with MAML, as opposed to the Finetune baseline, and note that this norm does not reflect generalization.

\section{Regularizing MAML}
\label{section: algo}

\begin{figure}[h!]
  \centering
  \includegraphics[width=0.9\linewidth]{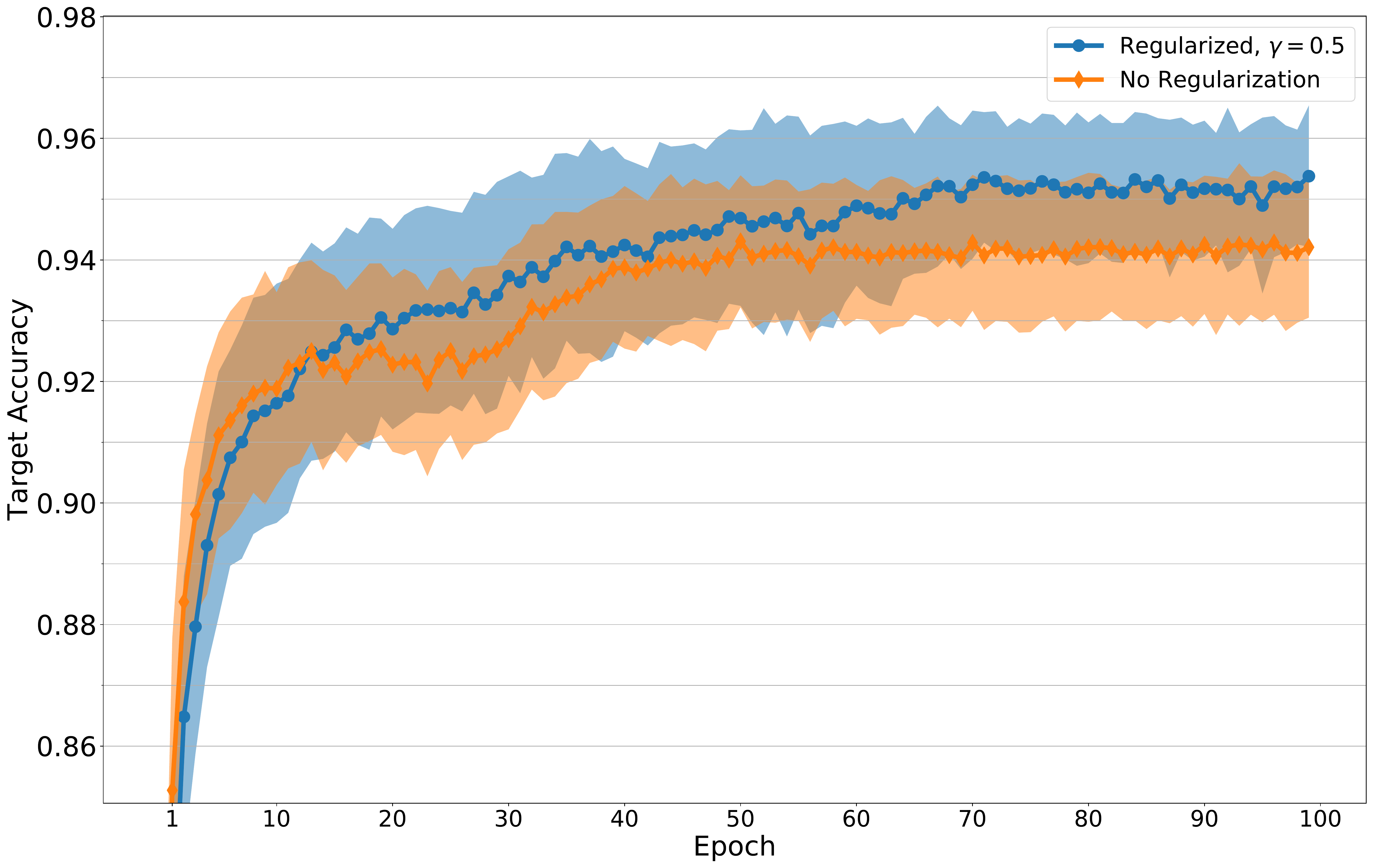}
  \caption{Average target accuracy on meta-test tasks using our proposed regularizer on MAML, for Omniglot 20-way 1-shot learning, with regularization coefficient $\gamma= 0.5$}
  \label{fig:regularization_experiments:omni_20w1s_1_step_angles}
\end{figure}

Based on our observations on the coherence of adaptation trajectories, we propose a modification of the MAML algorithm by adding a regularization term based on $\mathbb{E} [ \vec{\theta}_i^{\;\; T } \vec{\theta}_j ]$ . Within a meta-training iteration, we first let $f$ adapt to the $n$ training tasks $\mathcal{T}_i$ following Eq \ref{eq:maml:inner_loop}. We then compute the average direction vector $\vec{\theta}_{\mu} = \frac{1}{n} \sum_{i=1}^n \vec{\theta}_i$. From this point, we consider $\vec{\theta}_{\mu}$ to be fixed, such that $\nabla_{\theta}\vec{\theta}_{\mu} = 0$. For each task, we want to reduce the angle defined by $\vec{\theta}_i^{\;\; T } \vec{\theta}_{\mu}$, and thus introduce the penalty on $\Omega(\theta) = -\vec{\theta}_i^{\;\; T } \vec{\theta}_{\mu}$, obtaining the regularized solutions $\hat{\theta}_i$. The outer loop gradients are then computed, just like in MAML following Eq \ref{eq:maml:outer_loop}, but using these regularized solutions $\hat{\theta}_i$ instead of $\tilde{\theta}_i$. Note that after adding the regularizer, we consider it constant to avoid additional gradient computation overhead. We obtain the variant of MAML with regularized inner loop updates, as detailed in Algorithm \ref{alg:Regularized_MAML}:

\begin{algorithm}[t]
During a meta-training iteration:
\begin{algorithmic}[1]
    \STATE Sample a batch of $n$ tasks $\mathcal{T}_{i} \sim p(\mathcal{T})$
	\FORALL{$\mathcal{T}_i$}
        \STATE Perform the inner loop adaptation as in Eq. \ref{eq:maml:inner_loop}: $\tilde{\theta}_i = \theta^{s} - \alpha \sum_{t=0}^{T-1} \nabla_{\theta} \mathcal{L}(f(\mathcal{D}_i; \theta_i^{(t)}))$
	\ENDFOR
	\STATE Compute the average direction vector: $\vec{\theta}_{\mu} = \frac{1}{n} \sum_{i=1}^n \vec{\theta}_i$
	\\ Compute the corrected inner loop updates: \noindent
	\FORALL{$\mathcal{T}_i$}
	   \STATE $\hat{\theta}_i = \tilde{\theta}_i - \gamma \nabla_{\theta} \Omega(\theta)$, where $\Omega(\theta) = -\vec{\theta}_i^{\;\; T } \vec{\theta}_{\mu}$
	\ENDFOR
	\STATE Perform the meta-update as in Eq. \ref{eq:maml:outer_loop},  but using the corrected solutions:\\ $\theta^{s+1} = \theta^{s} - \beta \frac{1}{n} \sum_{i=1}^{n} \nabla_{\theta} \mathcal{L}(f(\mathcal{D}_i'; \hat{\theta}_i))$
\end{algorithmic}
\caption{Regularized MAML: Added penalty on angles between inner loop updates}
\label{alg:Regularized_MAML}
\end{algorithm}

We used this regularizer with MAML (Second-Order), for "Omniglot 20-way 1-shot", thereby tackling the most challenging few-shot classification setting for Omniglot. As shown in Figure \ref{fig:regularization_experiments:omni_20w1s_1_step_angles}, we observed an increase in meta-test target accuracy: the performance increases from $94.05\%$ to $95.38\%$ (average over five trials, 600 test tasks each), providing $\sim 23\%$ relative reduction in meta-test target error.

\section{Conclusion}

We experimentally demonstrate that when using gradient-based meta-learning algorithms such as MAML, meta-test solutions, obtained after adapting neural networks to new tasks via few-shot learning, become flatter, lower in loss, and further away from the meta-train solution, as meta-training progresses. We also show that those meta-test solutions keep getting flatter even when generalization starts to degrade, thus providing an experimental argument against the correlation between generalization and flat minima. More importantly, we empirically show that generalization to new tasks is correlated with the coherence between their adaptation trajectories, measured by the average cosine similarity between the adaptation trajectory directions, but also correlated with the coherence between the meta-test gradients, measured by the average inner product between meta-test gradient vectors evaluated at meta-train solution. Based on these observations, we propose a novel regularizer for MAML. As a future work, we plan to test the effectiveness of this regularizer on various datasets and meta-learning domains, architectures and gradient-based meta-learning algorithms.

\newpage

\bibliography{references}
\bibliographystyle{apalike}

\newpage
\appendix

\section{Additional Experimental Details}\label{appendix:experimental_details}

\subsection{Model Architectures}\label{appendix:experimental_details:model_architectures}

We use the architecture proposed by \citep{DBLP:journals/corr/VinyalsBLKW16} which is used by \citep{DBLP:journals/corr/FinnAL17}, consisting of 4 modules stacked on each other, each being composed of 64 filters of of 3 $\times$ 3 convolution, followed by a batch normalization layer, a ReLU activation layer, and a 2 $\times$ 2 max-pooling layer. With Omniglot, strided convolution is used instead of max-pooling, and images are downsampled to 28 $\times$ 28. With MiniImagenet, we used fewer filters to reduce overfitting, but used 48 while MAML used 32. As a loss function to minimize, we use cross-entropy between the predicted classes and the target classes.

\subsection{Meta-Learning datasets}\label{appendix:experimental_details:meta-learning_datasets}

The Omniglot dataset consists of a total of 1623 classes, each comprising 20 instances. The classes correspond to distinct characters, taken from 50 different datasets, but the taxonomy among characters isn't used. The MiniImagenet dataset comprises 64 training classes, 12 validation classes and 24 test classes. Each of those classes was randomly sampled from the original Imagenet dataset, and each contains 600 instances with a reduced size of 84 $\times$ 84.

\subsection{Hyperparameters used in meta-training and meta-testing}\label{appendix:experimental_details:meta-learning_hyperparameters}

We follow the same experimental setup as \citep{DBLP:journals/corr/FinnAL17} for training and testing the models using MAML and First-Order MAML. During meta-training, the inner loop updates are performed via five steps of full batch gradient descent (except for Section \ref{section:coherence_meta-test_gradients} where $T=1$), with a fixed learning rate $\alpha$ of $0.1$ for Omniglot and $0.01$ for MiniImagenet, while ADAM is used as the optimizer for the meta-update, without any learning rate scheduling, using a meta-learning rate $\beta$ of $0.001$. At meta-test time, adaptation to meta-test task is always performed by performing the same number of steps as for the meta-training inner loop updates. We use a mini-batch of 16 and 8 tasks for the 1-shot and 5-shot settings respectively, while for the MiniImagenet experiments, we use batches of 4 and 2 tasks for the 1-shot and 5-shots settings respectively. Let's also precise that, in \textit{k-shot} learning for an \textit{m-way} classification task $\mathcal{T}_i$, the set of support samples $\mathcal{D}_i$ comprises $k \times m$ samples. Each meta-training epoch comprises 500 meta-training iterations. 

For the finetuning baseline, we kept the same hyperparameters for the ADAM optimizer during meta-training, and for the adaptation during meta-test. We searched the training hyperparameter values for the mini-batch size and the number of iterations per epoch. Experiments are run for a 100 epochs each. In order to limit meta-overfitting and maximize the highest average meta-test target accuracy, the finetuning models see roughly 100 times less training data per epoch compared to a MAML training epoch. In order to evaluate the baseline on the 1-shot and 5-shot meta-test tasks, during training we used mini-batches of 64 images with 25 iterations per epoch for 1-shot learning, and mini-batches of 128 images with 12 iterations per epoch, for 5-shot learning. At meta-test time, we use Xavier initialization \citep{xavier} to initialize the weights of the final layer.

\section{Additional Experimental Results}

\subsection{Performance of models trained with MAML and First-Order MAML, on the few-shot learning settings}\label{appendix:experimental_results:maml_performance}
\begin{figure*}[h!]
\centering
    \subfloat[Meta-Train Accuracy]{%
        \includegraphics[width=0.49\linewidth]{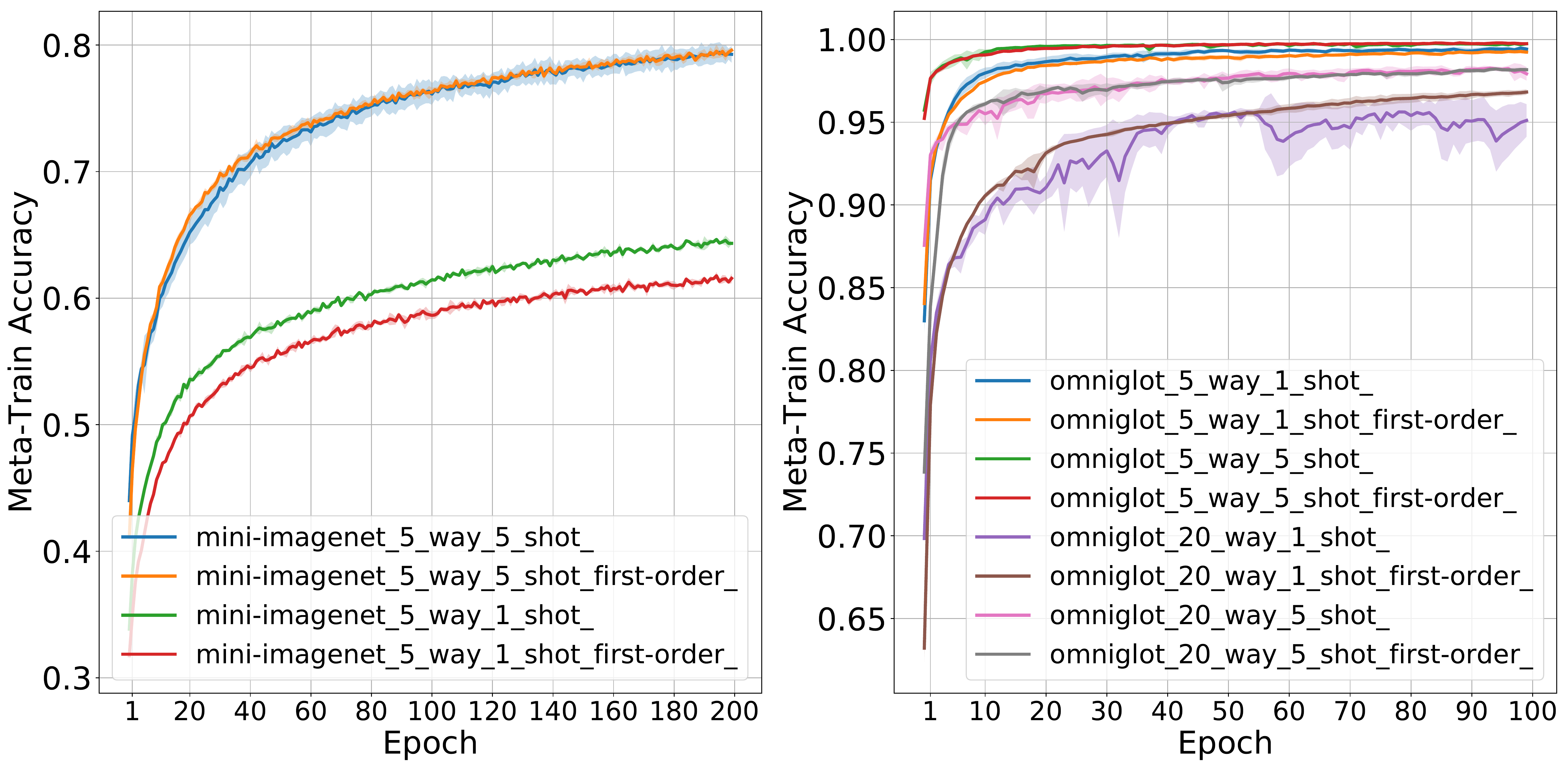}
        \label{fig:accuracies:meta-train}
    }
    \subfloat[Meta-Test Accuracy]{%
        \includegraphics[width=0.49\linewidth]{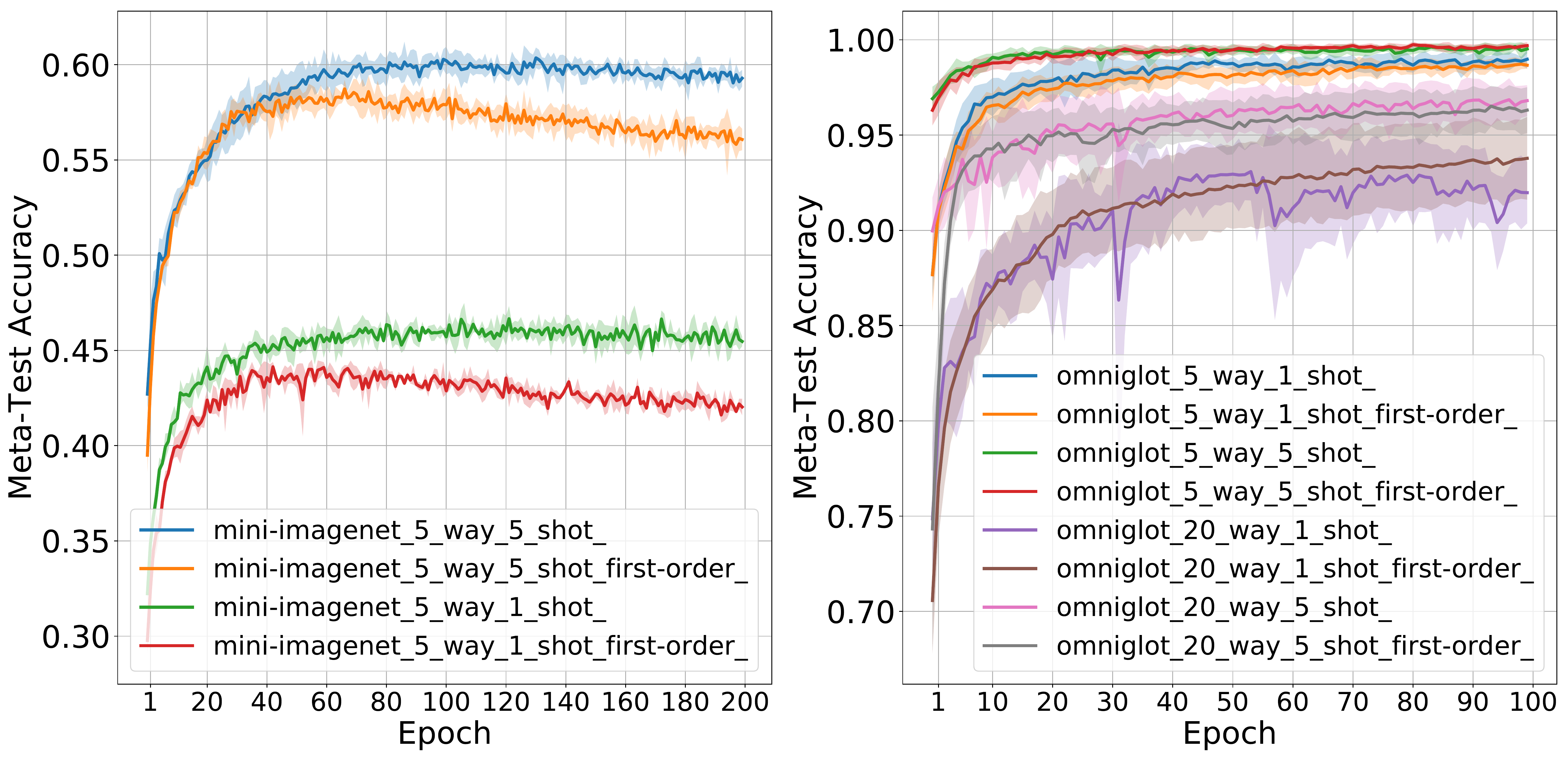}
        \label{fig:accuracies:meta-test}
    }
\caption{MAML: Accuracies on training and testing tasks}
\label{fig:accuracies}
\end{figure*}

\subsection{Coherence of adaptation trajectories}
\label{appen:angle}
\begin{figure*}[]
\centering
     \subfloat[MiniImagenet, 5-way, 1-shot, First-Order]{%
        \includegraphics[width=0.49\linewidth]{figures/angles/angles_mini-5w1s-fo.pdf}
        \label{fig:angles:mini_5w1s_fo}
    }
    \subfloat[MiniImagenet, 5-way, 1-shot, Second-Order]{%
        \includegraphics[width=0.49\linewidth]{figures/angles/angles_mini-5w1s.pdf}
        \label{fig:angles:mini_5w1s}
    }
    \newline
    \subfloat[Omniglot, 5-way, 5-shot, Second-Order]{%
        \includegraphics[width=0.49\linewidth]{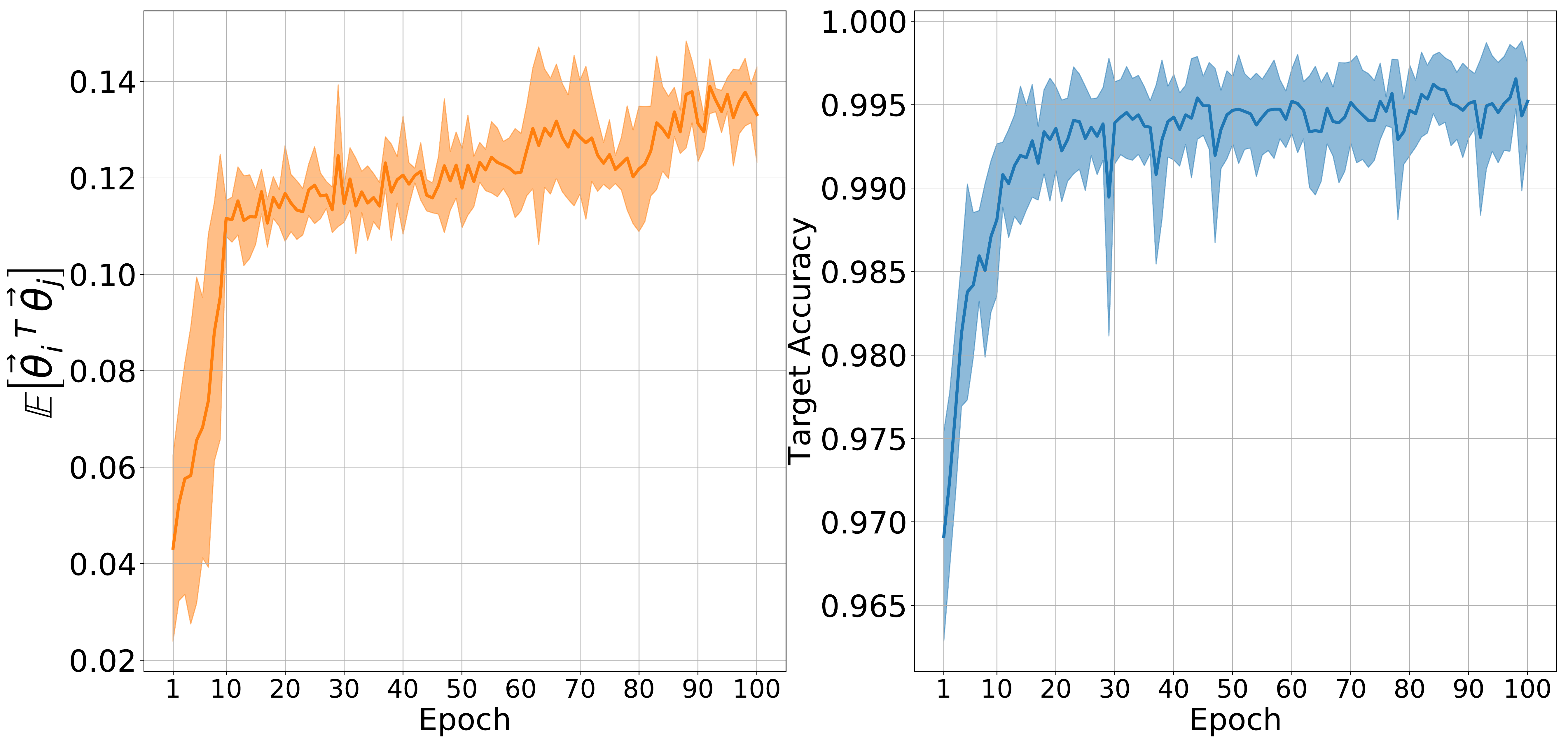}
        \label{fig:angles:omni_5w5s}
    }
    \subfloat[Omniglot, 20-way, 5-shot, Second-Order]{%
        \includegraphics[width=0.49\linewidth]{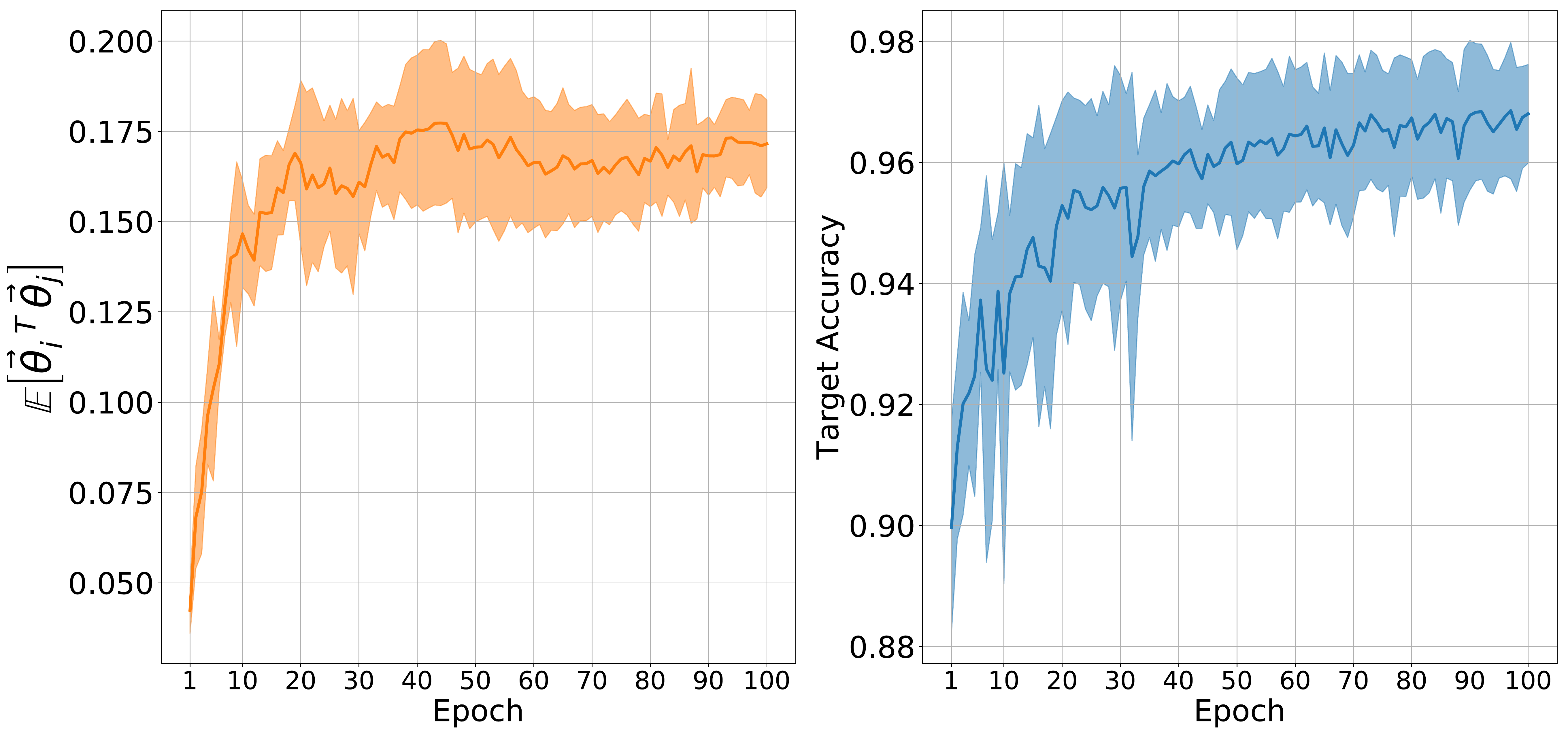}
        \label{fig:angles:omni_20w5s}
        }
\caption{Comparison between average inner product between trajectory directions and average target accuracy on meta-test tasks. Full set of experiments.}
\label{fig:appen:results:angles_vs_gen}
\end{figure*}

The relation between target accuracy on meta-test tasks, and angles between trajectory directions is presented in Figure \ref{fig:appen:results:angles_vs_gen}.

\subsection{Average inner product between meta-test gradients}
\label{appen:scalarprod}
\begin{figure*}[]
\centering
    \centering
    \subfloat[MiniImagenet, 5-way, 5-shot, First-Order]{%
        \includegraphics[width=0.49\linewidth]{figures/inner_products/inner_products_mini-imagenet_5_way_5_shot_first-order.pdf}
    }
    \subfloat[MiniImagenet, 5-way, 5-shot, Second-Order]{%
        \includegraphics[width=0.49\linewidth]{figures/inner_products/inner_products_mini-imagenet_5_way_5_shot.pdf}
        }
    \newline
    \subfloat[MiniImagenet, 5-way, 1-shot, First-Order]{%
        \includegraphics[width=0.49\linewidth]{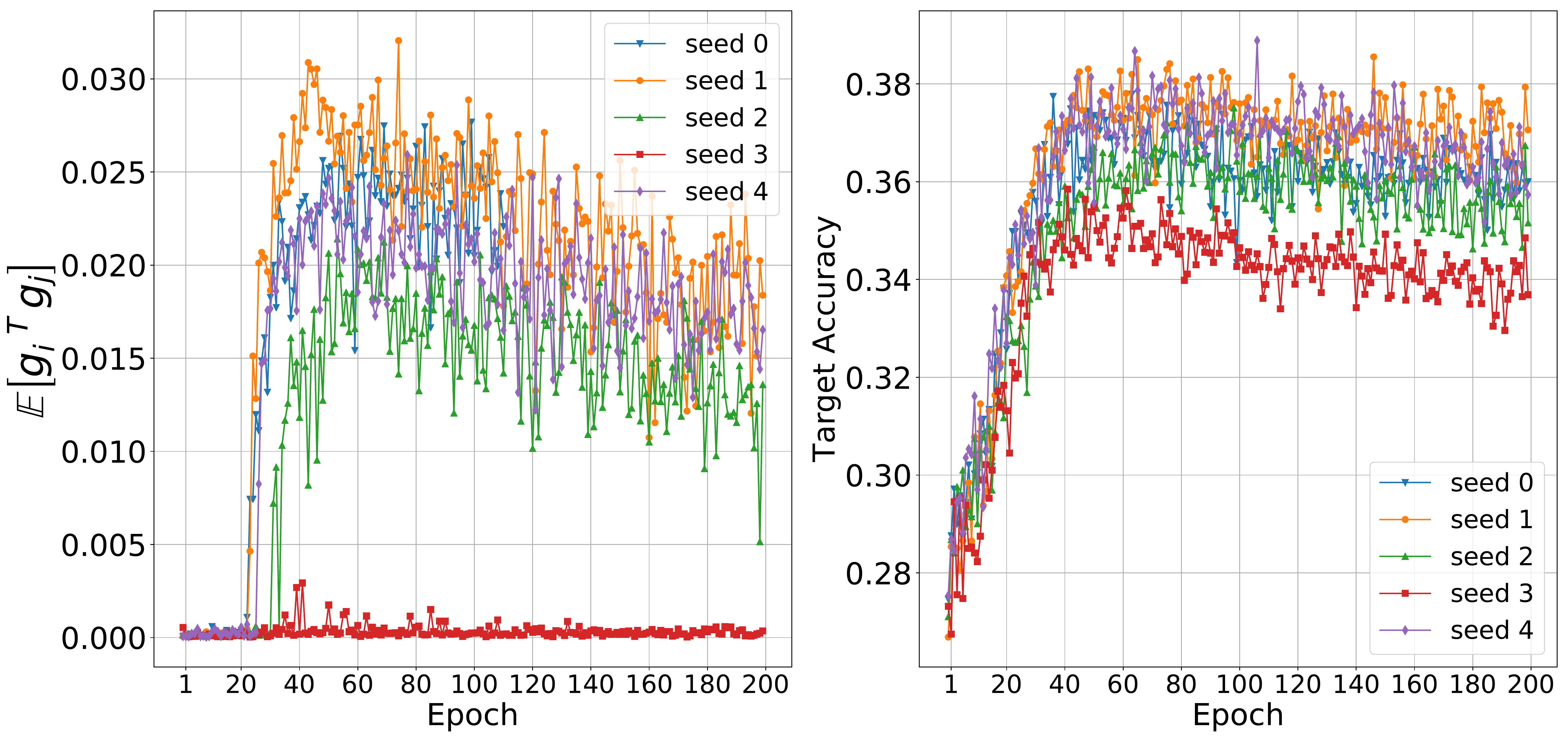}
    }
    \subfloat[MiniImagenet, 5-way, 1-shot, Second-Order]{%
        \includegraphics[width=0.49\linewidth]{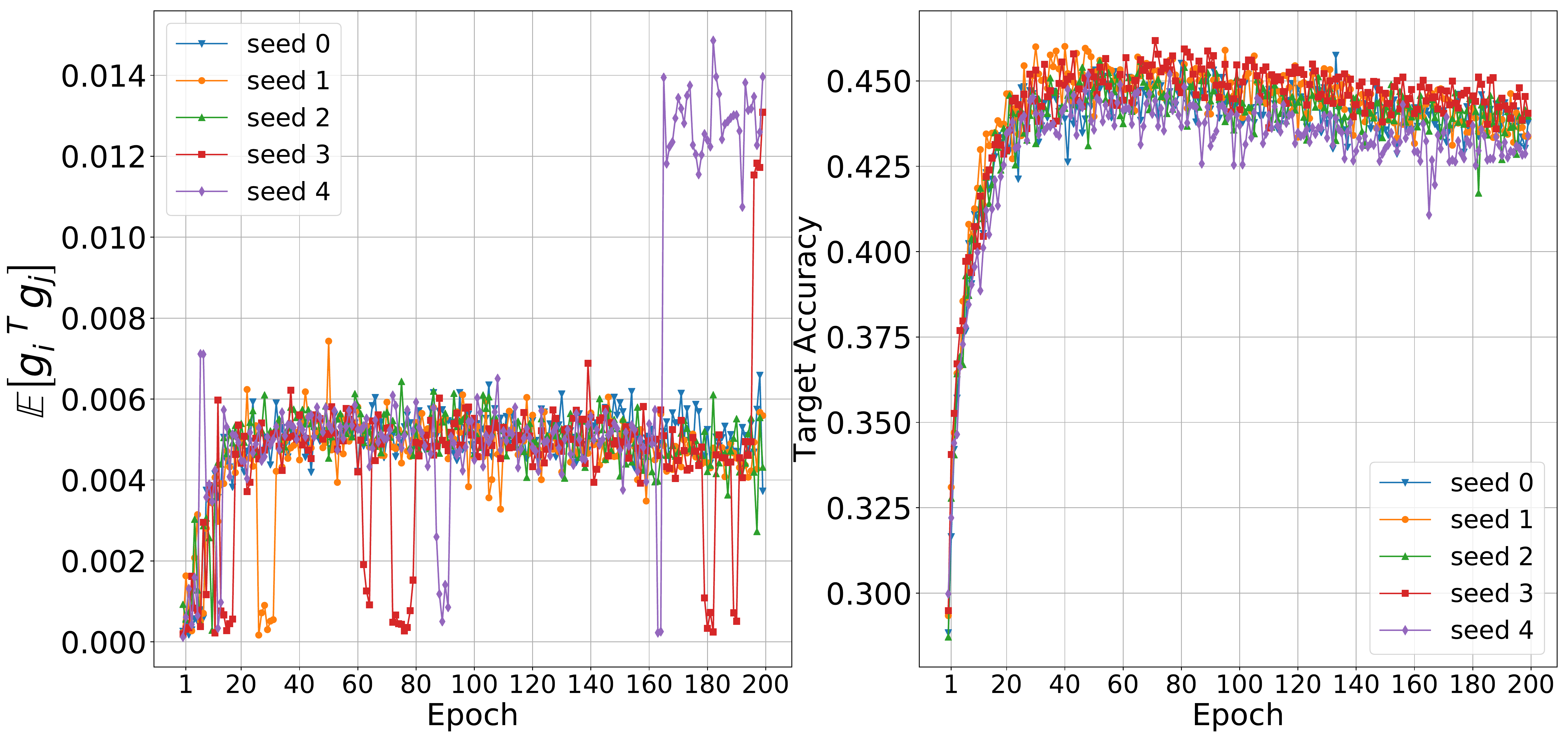}
    }
    \newline
    \subfloat[Omniglot, 20-way, 1-shot, Second-Order]{%
        \includegraphics[width=0.49\linewidth]{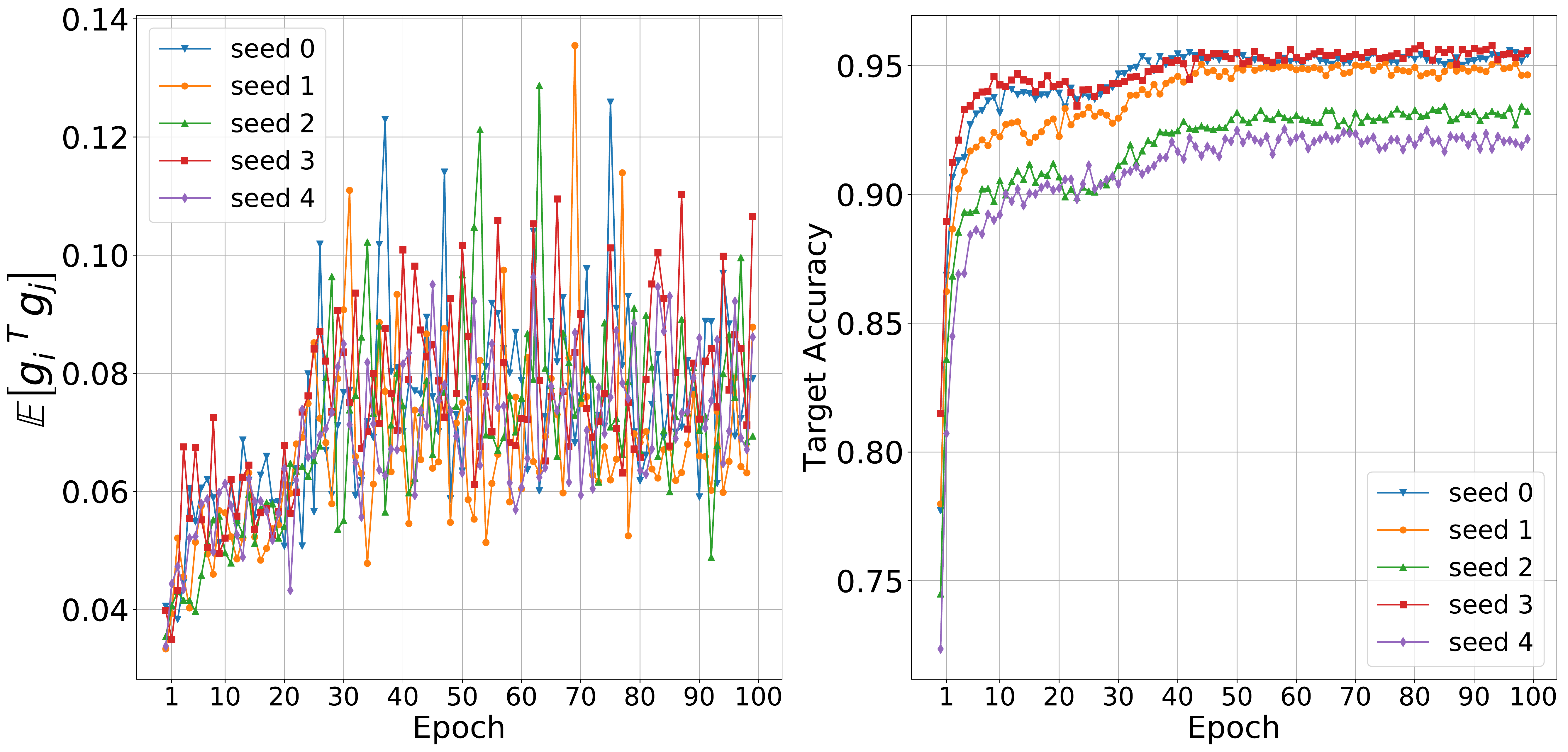}
    }
    \subfloat[Omniglot, 20-way, 5-shot, Second-Order]{%
        \includegraphics[width=0.49\linewidth]{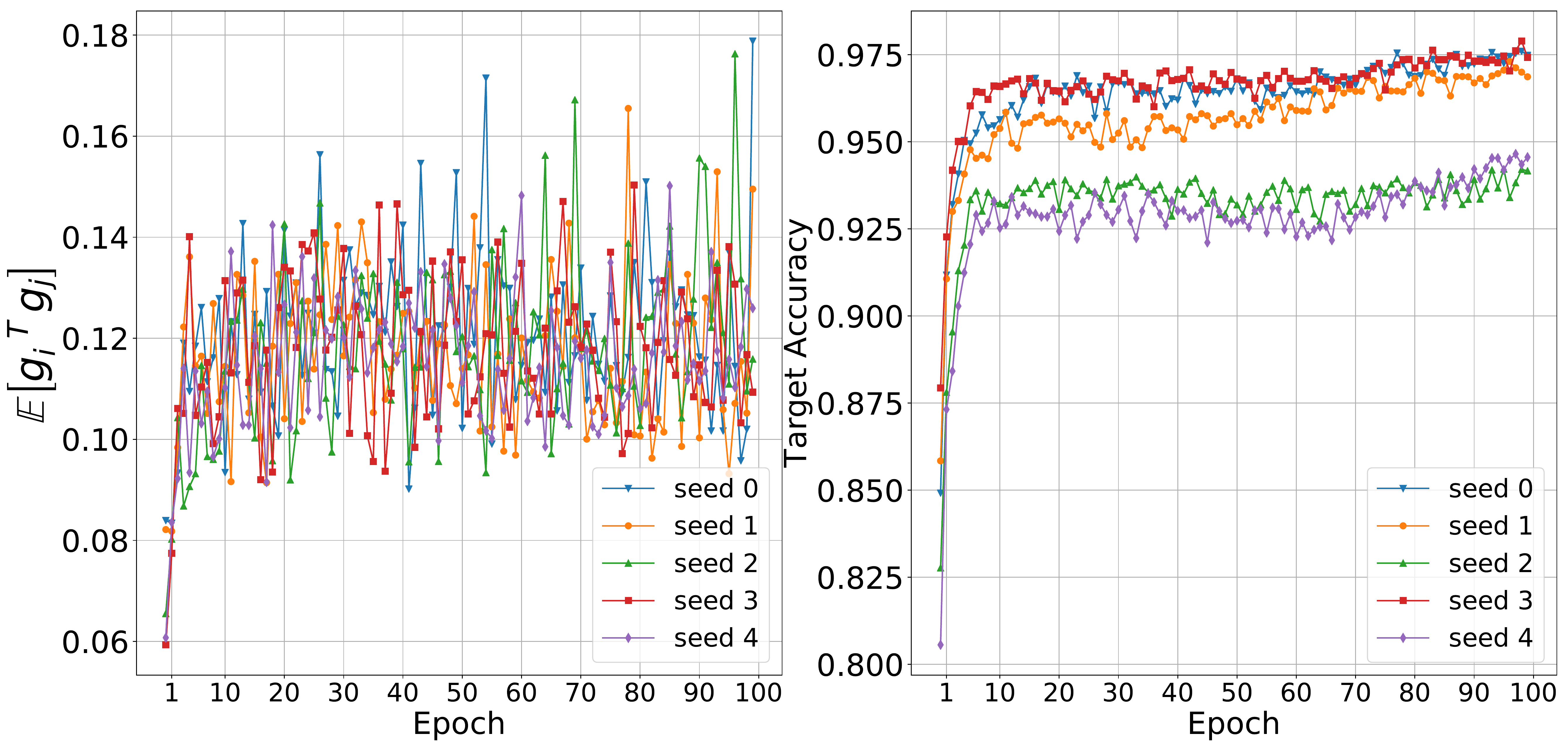}
        }        
\caption{Comparison between average inner product between trajectory displacement vectors, and average target accuracy on meta-test tasks. Full set of experiments.}
\label{fig:appen:results:dotprods_vs_gen}
\end{figure*}

The relation between target accuracy on meta-test tasks, and average inner product between meta-test gradients evaluated at meta-train solution, is presented in Figure \ref{fig:appen:results:dotprods_vs_gen}.

\end{document}